\documentclass{article}
\usepackage{a4wide}
\usepackage{graphicx}
\usepackage{booktabs,array,multirow,xcolor,colortbl}

\usepackage{apacite}
\usepackage[authoryear,round]{natbib}
\usepackage{nicefrac}
\usepackage{hyperref}

\begin{document}
\title{Computational historical linguistics}
\author{Gerhard J\"ager\\
University of T\"ubingen\\
Institute of Linguistics\\
Wilhelmstr.\ 19, 72074 T\"ubingen, Germany\\
Email: gerhard.jaeger@uni-tuebingen.de}
\date{}
\maketitle{}

\begin{abstract}
  Computational approaches to historical linguistics have been proposed since half a
  century. Within the last decade, this line of research has received a major boost, owing
  both to the transfer of ideas and software from computational biology and to the release
  of several large electronic data resources suitable for systematic comparative work.

  In this article, some of the central research topic of this new wave of computational
  historical linguistics are introduced and discussed. These are \emph{automatic
    assessment of genetic relatedness}, \emph{automatic cognate detection},
  \emph{phylogenetic inference} and \emph{ancestral state reconstruction}. They will be
  demonstrated by means of a case study of automatically reconstructing a Proto-Romance
  word list from lexical data of 50 modern Romance languages and dialects.
\end{abstract}
\newpage{}

\section{Introduction}

Historical linguistics is the oldest sub-discipline of linguistics, and it constitutes an
amazing success story. It gave us a clear idea of the laws governing language change, as
well as detailed insights into the languages --- and thus the cultures and living
conditions --- of prehistoric populations which left no written records. The diachronic
dimension of languages is essential for a proper understanding of their synchronic
properties. Also, the findings from historical linguistics are an important source of
information for other fields of prehistory studies, such as archaeology, paleoanthropology
and, in recent years, paleogenetics \citep[and many
others]{renfrew1987,pietrusewsky2008,anthony2010,haaketal2015}.

\begin{figure}
  \centering
  \includegraphics[width=.5\linewidth]{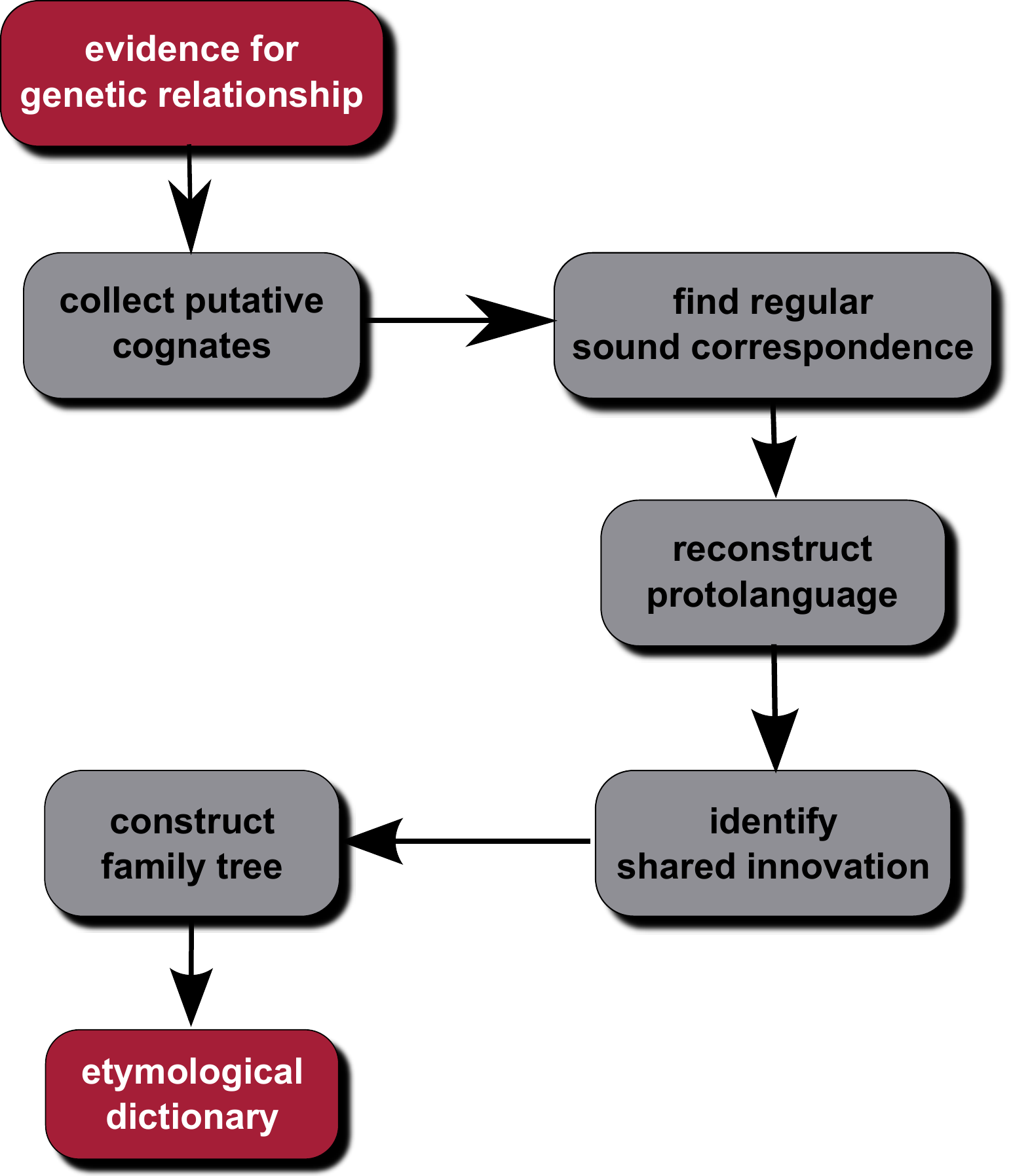}
  \caption{Workflow of the comparative method (according to \citealp{rossDurie96})}
  \label{fig:1}
\end{figure}

The success of historical linguistics is owed to a large degree to a collection of very
stringent methodological principles that go by the name of the \emph{comparative method}
\citep{meillet1925,weiss2015}. It can be summarized by the following workflow (from
\citealp[pp.\ 6--7]{rossDurie96}):
\begin{quote}
  \begin{enumerate}
  \item Determine on the strength of diagnostic evidence that a set of languages are
    genetically related, that is, that they constitute a `family'.
  \item Collect putative cognate sets for the family (both morphological paradigms and
    lexical items).
  \item Work out the sound correspondences from the cognate sets, putting `irregular'
    cognate sets on one side.
  \item Reconstruct the protolanguage of the family as follows:
    \begin{enumerate}
    \item [a.] Reconstruct the protophonology from the sound correspondences worked out in
      (3), using conventional wisdom regarding the directions of sound changes.
    \item [b.] Reconstruct protomorphemes (both morphological paradigms and lexical items)
      from the cognate sets collected in (2), using the protophonology reconstructed in
      (4a).
    \end{enumerate}
  \item Establish innovations (phonological, lexical, semantic, morphological,
    morphosyntactic) shared by groups of languages within the family relative to the
    reconstructed protolanguage.
  \item Tabulate the innovations established in (5) to arrive at an internal
    classification of the family, a `family tree'.
  \item Construct an etymological dictionary, tracing borrowings, semantic change, and so
    forth, for the lexicon of the family (or of one language of the family).
  \end{enumerate}
\end{quote}
In practice it is not applied in a linear, pipeline-like fashion. Rather, the results of
each intermediate step are subsequently used to inform earlier as well as later steps.
This workflows is graphically depicted in Figure \ref{fig:1}.

The steps (2)--(7) each involve a systematic, almost mechanical comparison and evaluation
of many options such as cognacy relations, proto-form reconstructions, or family
trees. The first step, establishing genetic relatedness, is less regimented, but it
generally involves a systematic comparison of many variables from multiple languages as
well. It is therefore not surprising that there have been many efforts to formalize parts
of this workflow to a degree sufficient to implement it on a computer.

Lexicostatistics (e.g.\ \citealp{swadesh52,swadesh55} and much subsequent work) can be
seen as an early attempt to give an algorithmic treatment of step (6), even though it
predates the computer age. Since the 1960s, several scholars applied computational methods
within the overall framework of lexicostatistics (cf.\ e.g.\ \citealp{embleton1986},
\emph{inter alia}). Likewise, there have been repeated efforts for computational
treatments of other aspects of the comparative method, such as
\citep{ringe1992,baxterManasterRamer2000,kessler2001} for step (1), \citep{kay1964} for
step (2), \citep{kondrak02} for steps (2) and (3), \citep{loweMazaudon1994} for steps (2)
and (4), \citep{oakes2000} for steps (2)--(7), \citep{covington96} for step (3), and
\citep{ringeWarnowTaylor2002} for step (6), to mention just a few of the earlier
contributions.

There is also a plethora of exciting work using historical corpora from different stages
of the same language to track lexical, grammatical and semantic change by computational
means (see for instance the overview in \citealp{hilpertGries2016} and the literature
cited therein).

While the mentioned proposals mostly constitute isolated efforts of historical and
computational linguists, the emerging field of computational historical linguistics
received a major impetus since the early 2000s by the work of computational biologists
such as Alexandre Bouchard-Côté, Russell Gray, Robert McMahon, Mark Pagel, or Tandy Warnow
and co-workers, who applied methods from their field to the problem of the reconstruction
language history, often in collaboration with linguists. This research trend might be
dubbed \emph{computational phylogenetic linguistics} as it heavily draws on techniques of
\emph{phylogenetic inference} from computational biology \citep{grayJordan2000,
  grayAtkinson03, mcmahonMcmahon2005, pageletal2007, atkinsonetal2008,
  grayDrummondGreenhill09, dunnetal11, bouckaertetal12, bouchardetal13, pagel2013,
  hruschkaetal15}.

In recent years, more and more large collections of comparative linguistic data become
available in digital form, giving the field another boost. The following list gives a
sample of the most commonly used databases; it is necessarily incomplete as new data
sources are continuously made public.
\begin{itemize}
\item \textbf{Cognate-coded word lists}
  \begin{itemize}
  \item \emph{Indo-European Lexical Cognacy Database} (IELex; \url{ielex.mpi.nl}):
    collection of 225-concept Swadesh lists from 163 Indo-European languages (based on
    \citealp{kruskalDyenBlack92}). Entries are given in orthography with manually assigned
    cognate classes; for part of the entries, IPA transcriptions are given.
  \item \emph{Austronesian Basic Vocabulary Database} (ABVD; \citealp{abvd};
    \url{language.psy.auckland.ac.nz/austronesian}): collection of 210-item Swadesh lists
    for 1,467 languages from the Pacific region, mostly belonging to the Austronesian
    language family. Entries are given in phonetic transcription with manually assigned
    cognate classes.
  \end{itemize}
\item \textbf{Phonetically transcribed word lists}
  \begin{itemize}
  \item \emph{ASJP database} (compiled by the \emph{Automatic Similarity Judgment
      Program}; \citealp{asjp17}; \url{asjp.clld.org}): collection of word lists
    for 7,221 doculects (languages and dialects) over 40 concepts (100-item word lists for ca.\ 300
    languages); entries are given in phonetic transcription.
  \end{itemize}
\item \textbf{Grammatical and typological classifications}
  \begin{itemize}
  \item \emph{World Atlas of Language Structure} (\citealp{wals};
    \url{wals.info}): manual expert classifications of 2,679 languages and
    dialects according to 192 typological features.
  \item \emph{Syntactic Structures of the World's Languages}
    (\url{sswl.railsplayground.net}): Classification of 274 languages according to
    148 syntactic features.
  \end{itemize}
\item \textbf{Expert language classifications}
  \begin{itemize}
  \item \emph{Ethnologue} (\citealp{ethnologue2016}; \url{https://www.ethnologue.com}):
    genetic classification of 7,457 languages, alongside with information about number of
    speakers, location, and viability.
  \item \emph{Glottolog} (\citealp{glottolog2_7}; \url{glottolog.org}): genetic
    classification of 7,943 languages and dialects, alongside with information about
    geographic locations and extensive bibliographic references
  \end{itemize}
\end{itemize}

Additionally there is a growing body of diachronic corpora of various languages. The focus
of this article is on computational work inspired by the comparative method, so this line
of work will not further be covered here.

\section{A program for \emph{computational historical linguistics}}

Conceived in a broad sense, computational historical linguistics comprises all efforts
deploying computational methods to answer questions about the history of natural
languages. As spelled out above, there is a decade-old tradition of this kind of research.

In this article, however, the term will be used in a rather narrow sense to describe an
emerging subfield which has reached a certain degree of convergence regarding research
goals, suitable data source and computational methods and tools to be deployed. I will
used the abbreviation \emph{CHL} to refer to computational historical linguistics in this
narrow sense. The following remarks strive to describe this emerging consensus. They are
partially programmatic in nature though; not all researchers active in this domain will
agree with all of them.

CHL is informed by three intellectual traditions:
\begin{itemize}
\item the \textbf{comparative method} of classical historical linguistics,
\item \textbf{computational biology}, especially regarding \emph{sequence alignment}
  \citep[cf.][]{durbinetal98} and \emph{phylogenetic inference} \citep[see,
  e.g.,][]{ewansGrant,chenKuoLewis}, and
\item computational linguistics in general, especially modern statistical \textbf{Natural
    Language Processing} (NLP).
\end{itemize}

CHL shares, to a large degree, the research objectives of the \emph{comparative
  method}. The goal is to reconstruct the historical processes that led to the observed
diversity of extant or documented ancient languages. This involves, \emph{inter alia}
establishing cognacy relations between words and morphemes, identifying regular sound
correspondences, inferring family trees (\emph{phylogenetic trees} or simply
\emph{phylogenies} in the biology-inspired terminology common in CHL), reconstructing
proto-forms and historical processes such as sound laws and lexical innovations.

CHL's guiding model is adapted from \emph{computational biology}. The history of a group
of languages is represented by a phylogenetic tree (including branch lengths), with
observed linguistic varieties at the leafs of the tree. Splits in a tree represent
diversification events, i.e., the separation of an ancient language into
daughter-lineages. Language change is conceptualized as a continuous-time Markov process
applying to discrete, finite-values characters. (Details will be spelled out below.)
Inference amounts to finding the model (a phylogenetic tree plus a parameterization of the
Markov process) that best explains the observed data.

Last but not least, CHL adopts techniques and methodological guidelines from
\emph{statistical NLP}. The pertinent computational tools, such as string comparison
algorithms, to a certain degree overlap with those inspired by computational
biology. Equally important are certain methodological standards from NLP and machine
learning.

Generally, work in CHL is a kind of \emph{inference}, where a collection of data are used
as input (premises) to produce output data (conclusions). Input data can be phonetically
or orthographically transcribed word lists, pairwise or multiply aligned word lists,
grammatical feature vectors etc. Output data are for instance cognate class labels,
alignments, phylogenies, or proto-form reconstructions. Inference is performed by
constructing a \emph{model} and \emph{training} its parameters. Following the standards in
statistical NLP, the following guiding principles are desirable when performing inference:
\begin{itemize}
\item \textbf{Replicability.} All data used in a study, including all manual
  pre-processing steps, is available to the scientific community. Likewise, each
  computational inference step is either documented in sufficient detail to enable
  re-implementation, or made available as source code.
\item \textbf{Rigorous evaluation.} The quality of the inference, or \emph{goodness of
    fit} of the trained model, is evaluated by applying a well-defined quantitative
  measure to the output of the inference. This measure is applicable to competing model
  for the same inference task, facilitating model comparison and model selection.
\item \textbf{Separation of training and test data.} Different data sets are used for
  training and evaluating a model. 
\item \textbf{Only raw data as input.} Only such data are used as input for inference that
  can be obtained without making prior assumptions about the inference task. For instance,
  word lists in orthographic or phonetic transcription are suitable as input if the
  transcriptions were produced without using diachronic information.
\end{itemize}

The final criterion is perhaps the most contentious one. It excludes, for instance, the
use of orthographic information in languages such as English or French for training
purposes, as the orthographic conventions of those languages reflect the phonetics of
earlier stages. Also, it follows that the cognate class labels from databases such as
IELex or ABVD, as well as expert classifications such as Ethnologue or Glottolog, are
unsuitable as input for inference and should only be used as gold standard for training
and testing.

Conceived this way, CHL is much narrower in scope than, e.g., computational phylogenetic
linguistics. For instance, inference about the time depth and homeland of language
families \citep[such as][]{grayAtkinson03, bouckaertetal12} is hard to fit into this
framework as long as there are no independent test data to evaluate models against (but
see \citealp{rama13}). Also, it is common practice in computational phylogenetic
linguistics to use manually collected cognate classifications as input for inference
\citep{grayJordan2000, grayAtkinson03, pageletal2007, atkinsonetal2008,
  grayDrummondGreenhill09, dunnetal11, bouckaertetal12, bouchardetal13, pagel2013,
  hruschkaetal15}. While the results obtained this way are highly valuable and insightful,
they are not fully replicable, since expert cognacy judgments are necessarily subjective
and variable. Also, the methods used in the work mentioned do not generalize easily to
under-studied language families, since correctly identifying cognates between distantly
related languages requires the prior application of the classical comparative method, and
the necessary research has not been done with equal intensity for all language families.

\section{A case study: reconstructing Proto-Romance}

\label{sec:protoRomance}

In this section a case study will be presented that illustrates many of the techniques
common in current CHL. Training data are 40-item word lists from 50 Romance (excluding
Latin) and 3 Albanian\footnote{The inclusion of Albanian will be motivated below.}
languages and dialects in phonetic transcription from the ASJP database \citep{asjp17}
(version 17, accessed on August 2, 2016 from
\url{asjp.clld.org/static/download/asjp-dataset.tab.zip}). The inference goal is the
reconstruction of the corresponding word list from the latest common ancestor of the
Romance languages and dialects (Proto-Romance, i.e., some version of Vulgar Latin). The
results will be tested against the Latin word lists from ASJP. A subset of the data used
is shown in Table \ref{tab:1} for illustration.
\begin{table}
  \centering
  \begin{tabular}{>{\em } l|llll|l}
    \toprule
    \emph{concept} & ALBANIAN & SPANISH  & ITALIAN  & ROMANIAN & LATIN     \\
    \midrule
    horn           & bri      & kerno    & korno    & korn     & kornu     \\
    knee           & Tu       & rodiya   & jinokkyo & jenuNk   & genu      \\
    mountain       & mal      & sero     & monta5a  & munte    & mons      \\
    liver          & m3lCi    & igado    & fegato   & fikat    & yekur     \\
    we             & ne       & nosotros & noi      & noi      & nos       \\
    you            & ju       & ustet    & tu       & tu       & tu        \\
    person         & vet3     & persona  & persona  & persoan3 & persona   \\
    louse          & morr     & pioho    & pidokko  & p3duke   & pedikulus \\
    new            & iri      & nuevo    & nwovo    & nou      & nowus     \\
    hear           & d3gyoy   & oir      & ud       & auz      & audire    \\
    sun            & dyell    & sol      & sole     & soare    & sol       \\
    tree           & dru      & arbol    & albero   & pom      & arbor     \\
    breast         & kraharor & peCo     & pEtto    & pept     & pektus    \\
    drink          & pirye    & bebe     & bere     & bea      & bibere    \\
    hand           & dor3     & mano     & mano     & m3n3     & manus     \\
    die            & vdes     & mori     & mor      & mur      & mori      \\
    name           & em3r     & nombre   & nome     & nume     & nomen     \\
    eye            & si       & oho      & okkyo    & ok       & okulus    \\
    \bottomrule
  \end{tabular}
  \caption{Sample of word lists used}
  \label{tab:1}
\end{table}
The phonetic transcriptions use the 41 ASJP sound classes (cf.\
\citealp{brownetal13}). Diacritics are removed. If the database lists more than one
translations for a concept in a given language, only the first one is used.

The following steps will be performed (mirroring to a large degree the steps of the
comparative method):
\begin{enumerate}
\item Demonstrate that the Romance languages and dialects are related.
\item Compute pairwise string alignments and string similarities between synonymous words
  from different languages/dialects.
\item Cluster the words for each concept into automatically inferred cognate classes.
\item Infer a phylogenetic tree (or a collection of trees).
\item Perform Ancestral State Reconstruction for cognate classes to infer the cognate
  class of the Proto-Romance word for each concept.
\item Perform multiple sequence alignment of the reflexes of those cognate classes within
  the Romance languages and dialects.
\item Perform Ancestral State reconstruction to infer the state (sound class or gap) of
  each column in the multiple sequence alignments.
\item Compare the results to the Latin ASJP word list.
\end{enumerate}

\subsection{Demonstration of genetic relationship}

In \citep{jaeger13ldc} a dissimilarity measure between ASJP word lists is developed. Space
does not permit to explain it in any detail here. Suffice it to say that this measure is
based on the average string similarity between the corresponding elements of two word
lists while controlling for the possibility of chance similarities. Let us call this
dissimilarity measure between two word lists the \emph{PMI distance}, since it makes
crucial use of the \emph{pointwise mutual information} (PMI) between phonetic strings.

To demonstrate that all Romance languages and dialects used in this study are mutually
related, we will use the ASJP word lists from Papunesia, i.e., ``all islands between
Sumatra and the Americas, excluding islands off Australia and excluding Japan and islands
to the North of it'' \citep{glottolog2_7} as training data and the ASJP word lists from
Africa as test data.\footnote{We chose different macro-areas for training and testing to
  minimize the risk that the data are non-independent due to common ancestry or language
  contact.} Input for inference are PMI distances between pairs of languages/dialect, and
the output is the classification of this pair as \emph{related} or \emph{unrelated}, where
two doculects count as related if they belong to the same language family according to the
Glottolog classification.
\begin{figure}[t]
  \centering
  \includegraphics[width=.5\linewidth]{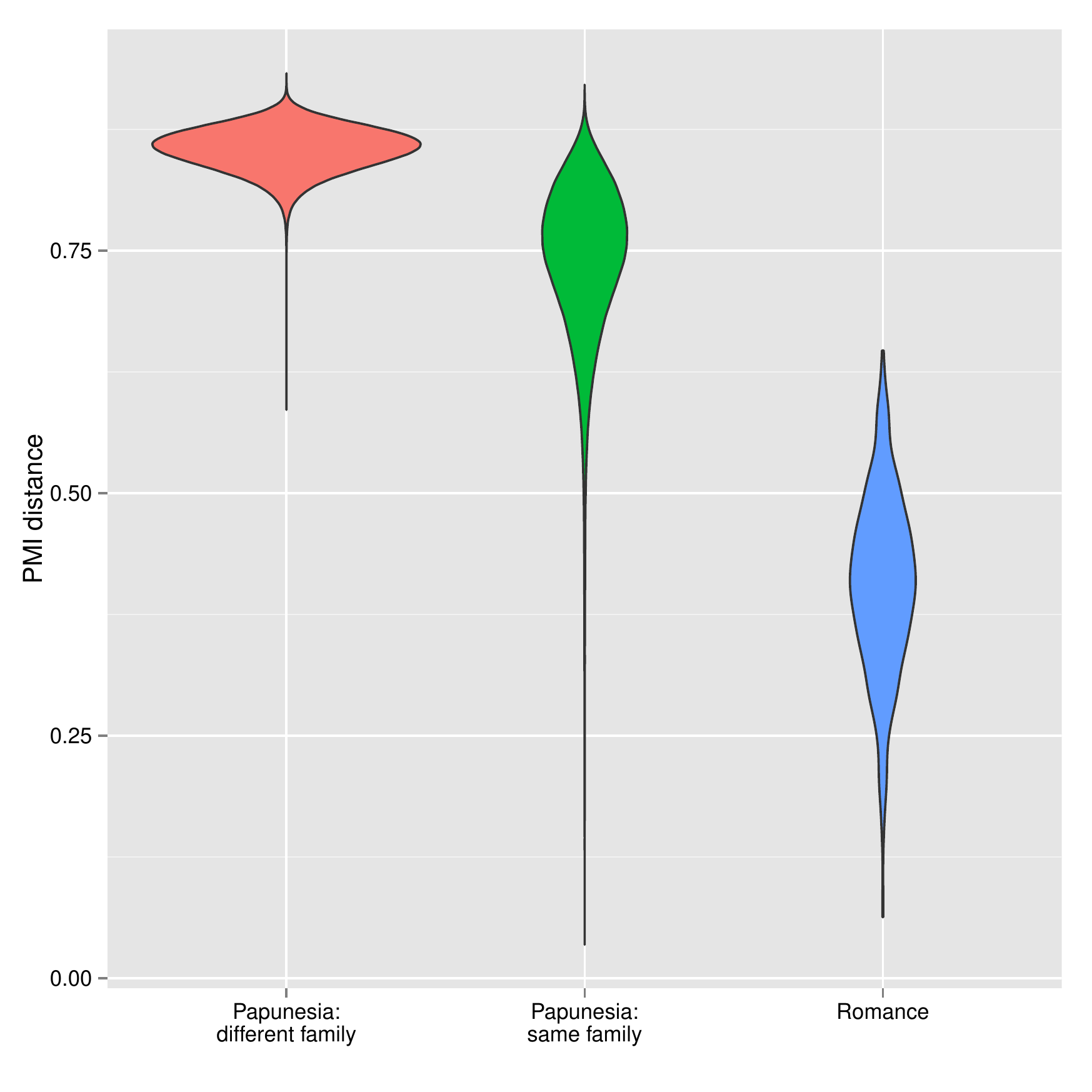}
  \caption{PMI distances between related and unrelated doculects from Papunesia, and
    between the Romance doculects}
  \label{fig:2}
\end{figure}
The graphics illustrates that all doculect pairs with a PMI distance $\leq 0.75$ are, with
a very high probability, related. The largest PMI distance among Romance dialects (between
Aromanian and Nones) is 0.65. 

A statistical test confirms this impression. We fitted a cumulative density estimation
for the PMI distances of the unrelated doculect pairs from the training data, using the
R-package \emph{logspline} \citep{logspline}. If a pair of doculects has a PMI distance
$d$, the value of the cumulative density function for $d$ can then be interpreted as the
(one-sided) $p$-value for the null hypothesis that the doculects are unrelated.

Using a threshold of $\alpha=0.0001$, we say that a doculect pair is predicted to be
related if the model predicts it to be unrelated with a probability $\leq \alpha$. In
Table \ref{tab:2}, the predictions are tabulated against the Glottolog gold standard.
\begin{table}
  \centering
  \begin{tabular}{lrr}
    \toprule
                          & \multicolumn{2}{c}{Glottolog:} \\
                          & unrelated & related            \\\midrule
    prediction: unrelated & 1,254,726 & 787,023            \\
    prediction: related   & 532       & 153,279            \\\bottomrule
  \end{tabular}
  \caption{Contingency table of gold standard versus prediction for the test data}
  \label{tab:2}
\end{table}
These results amount to ca.\ $0.3\%$ of false positives and ca.\ $84\%$ of false
negatives. This test ensures that the chosen model and threshold is sufficiently
conservative to keep the risk of wrongly assessing doculects to be related small. Since
the method so conservative, it produces a large amount of false negatives though.

In the next step, we compute the probability of all pairs of Romance doculects to be
unrelated, using the model obtained from the training data. Using the Holm-Bonferroni
method to control for multiple tests, the highest $p$-value for the null hypothesis that
the doculect pair in question is unrelated is $1.8\times 10^{-5}$, i.e., all adjusted
$p$-values are $<\alpha$. We can therefore reject the null hypothesis for all Romance
doculect pairs.

\subsection{Pairwise string comparison}
\label{sec:pairwise}

All subsequent steps rely on a systematic comparison of words for the same concept from
different doculects. Let us consider as an example the words for \emph{water} from Catalan
and Italian from ASJP, ``aigua'' and ``acqua''. Both are descendants of the Latin ``aqua''
\citep[p.\ 46]{meyerLuebke}.  In ASJP transcription, these are
\texttt{aixw\textasciitilde{}3} and \texttt{akwa}. The sequence
\texttt{w\textasciitilde{}} in the Catalan word encodes a diacritic (indicating
labialization of the preceding segment) and is removed in the subsequent processing steps.

A \emph{pairwise sequence alignment} of two strings arranges them in such a way that
corresponding segments are aligned, possibly inserting gap symbols for segments in one
string that have no correspondent in the other string. For the example, the historically
correct alignment would arguably be as follows:
\begin{center}
  \begin{tabular}{c}
    \texttt{aix-3}\\
    \texttt{a-kwa}
  \end{tabular}
\end{center}

In this study, the quality of a pairwise alignment is quantified as its aggregate
\emph{pointwise mutual information} (PMI). (See \citealp{list14} for a different
approach.) The PMI between two sound classes $a,b$ is defined as
\begin{eqnarray}
  \mathit{PMI}(a,b) &\doteq& \log\frac{s(a,b)}{q(a)q(b)}.\label{eq:1}
\end{eqnarray}
Here $s(a,b)$ is the probability that $a$ is aligned to $b$ in a correct alignment, and
$q(a), q(b)$ are the probabilities of occurrence of $a$ and $b$ in a string. If one of the
two symbols is a gap, the PMI score is a \emph{gap penalty}. We use \emph{affine gap
  penalties}, i.e., the gap penalty is reduced if the gap is preceded by another gap.

If the PMI scores for each pair of sound classes, and the gap penalties are known, the
best alignment between two strings (i.e., the alignment maximizing the aggregate PMI
score) can efficiently be computed using the Needleman-Wunsch algorithm
\citep{needlemanWunsch}.

The quantities $s(a,b)$ and $q(a), q(b)$ must be estimated from the data. Here we follow a
simplified version of the parameter estimation technique from \citep{jaeger13ldc}. In a
first step, we set
\[
  \mathit{PMI}_0(a,b) \doteq 
  \left\{
    \begin{array}[c]{rl}
      0 & \mbox{if }a=b\\
      -1& \mbox{else.}
    \end{array}
  \right.
\]
Also, we set the initial gap penalties to $-1$. (This amounts to \emph{Levenshtein
  alignment}.) Using these parameters, all pairs of word for the same concept from
different doculects are aligned.

From those alignments, $s(a,b)$ is estimated as the relative frequency of $a$ and $b$
being aligned among all non-gap alignment pairs, while $q(a)$ is estimated as the relative
frequency of sound class $a$ in the data. The PMI scores are then estimated using equation
(\ref{eq:1}). For the gap penalties we used the values from \citep{jaeger13ldc}, i.e.,
$-2.49$ for opening gaps and $-1.70$ for extending gaps. Using those parameters, all
synonymous word pairs are re-aligned. 

In the next step, only word pairs with an aggregate PMI score $\geq 4.45$ are used. (This
threshold is taken from \citealp{jaeger13ldc} as well.) Those word pairs are re-aligned
and the PMI scores are re-estimated. This step is repeated ten times. 

The threshold of $4.45$ is rather strict; almost all word pairs above this threshold are
either cognates or loans. For instance, for the language pair Italian/Albanian, the only
translation pair with a higher PMI score is Italian \texttt{peSe}/ Albanian \texttt{peSk}
(``fish''), where the former is a descendant and the latter a loan from Latin
\emph{piscis} (cf.\ \url{http://ielex.mpi.nl}). For Spanish/Romanian, two rather divergent
Romance languages, we find eight such word pairs. They are shown alongside with the
inferred alignments in Table \ref{tab:3}.

\begin{table}[h]
  \centering
  \begin{tabular}[c]{>{\it} l>{\tt} lr}
    \toprule
    \normalfont{}concept    & \normalfont{}alignment & PMI score                \\\midrule
    \multirow{2}{*}{person} & perso-na               & \multirow{2}{*}{$14.23$} \\
                            & persoan3               &                          \\\midrule
    \multirow{2}{*}{tooth}  & diente                 & \multirow{2}{*}{$10.13$} \\
                            & di-nte                 &                          \\\midrule
    \multirow{2}{*}{blood}  & sangre                 & \multirow{2}{*}{$8.04$}  \\
                            & s3nj-e                 &                          \\\midrule
    \multirow{2}{*}{hand}   & mano                   & \multirow{2}{*}{$6.71$}  \\
                            & m3n3                   &                          \\\midrule
    \multirow{2}{*}{one}    & uno                    & \multirow{2}{*}{$5.61$}  \\
                            & unu                    &                          \\\midrule
    \multirow{2}{*}{die}    & mori                   & \multirow{2}{*}{$5.16$}  \\
                            & mur-                   &                          \\\midrule
    \multirow{2}{*}{come}   & veni                   & \multirow{2}{*}{$5.01$}  \\
                            & ven-                   &                          \\\midrule
    \multirow{2}{*}{name}   & nombre                 & \multirow{2}{*}{$4.98$}  \\
                            & num--e                 &                          \\\midrule
  \end{tabular}
  \caption{Word pair alignments from Spanish and Romanian}
  \label{tab:3}
\end{table}

The aggregate PMI score for the best alignment between two strings is a measure for the
degree of similarity between the strings. We will call it the \emph{PMI similarity}
henceforth.

\subsection{Cognate clustering}
\label{sec:cc}

Automatic cognate detection is an area of active investigation in CHL \citep[\emph{inter
  alia}]{dolgopolsky86,Bergsma2007,hallKlein2010,Turchin2010,Hauer2011,list2012lexstat,list14,rama2015,jaegerSofroniev16Konvens,jaegerListSofroniev17}. For
the present study, we chose a rather simple approach based on unsupervised learning.

\begin{figure}[h]
  \centering
  \includegraphics[width=.5\linewidth]{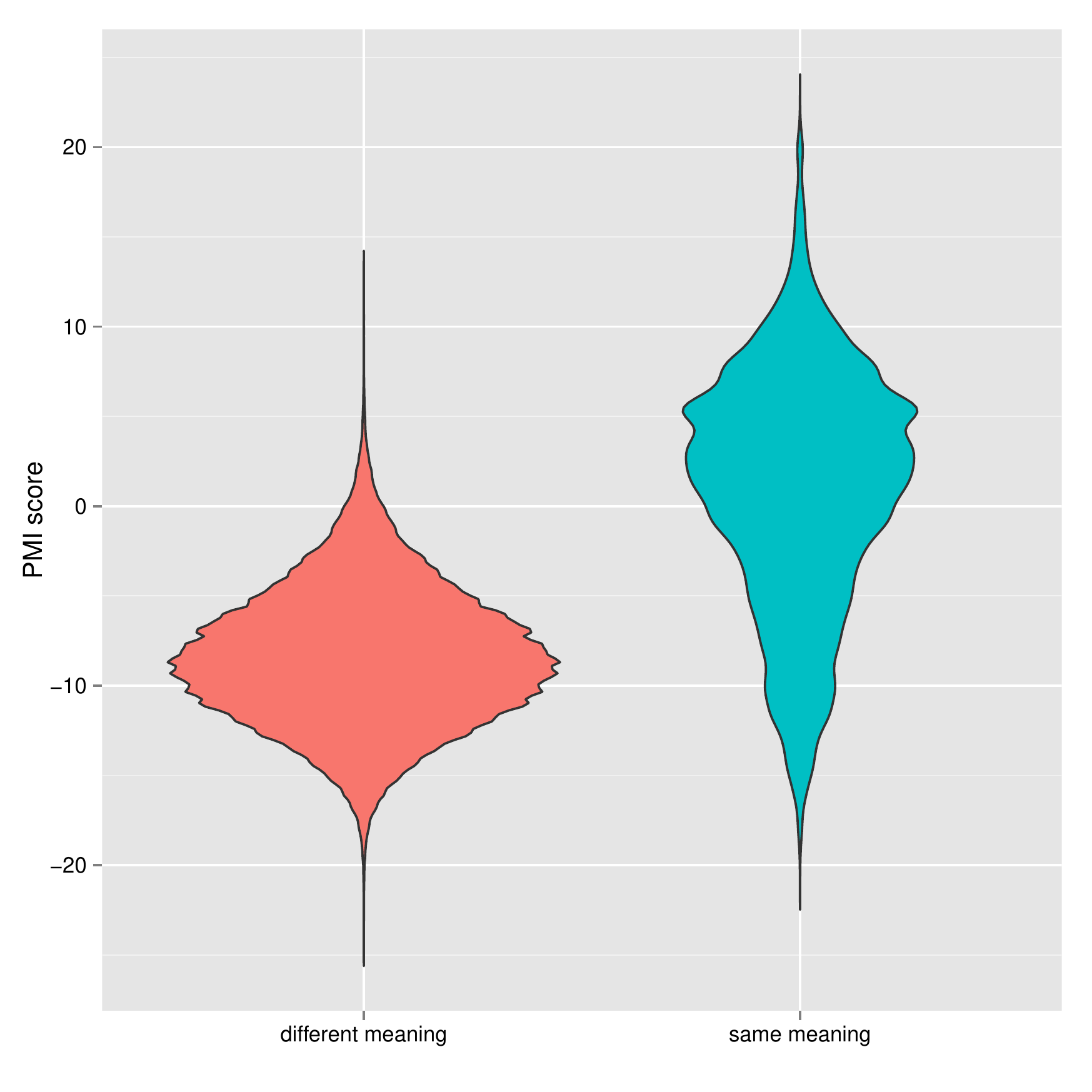}
  \caption{PMI similarities for synonymous and non-synonymous word pairs}
  \label{fig:3}
\end{figure}

Figure \ref{fig:3} shows the PMI similarities for words from different doculects have
different or identical meanings.
Within our data, synonymous word pairs are, on average, more similar to each other than
non-synonymous ones. The most plausible explanation for this effect is that the synonymous
word pairs contain a large proportion of cognate pairs. Therefore ``identity of meaning''
will be used as a proxy for ``being cognate''.

We fitted a logistic regression with PMI similarity as independent and synonymy as
dependent variable.

For each concept, a weighted graph is constructed, with the words denoting this concept as
vertices. Two vertices are connected if the predicted probability of these words to be
synonymous (based on their PMI similarity and the logistic regression model) is
$\geq 0.5$. The weight of each edge equals the predicted probabilities. The nodes of the
graph are clustered using the weighted version of the \emph{Label Propagation} algorithm
\citep{labelPropagation} as implemented in the \emph{igraph} software \citep{igraph}. As a
result, a \emph{class label} is assigned to each word. Non-synonymous words never carry
the same class label.\footnote{The implicit assumption underlying this procedure is that
  cognate words always have the same meaning. This is evidently false when considering the
  entire lexicon. There is a plethora of examples, such as as English \emph{deer} vs.\
  German \emph{Tier} ``animal'', which are cognate \citep[cf.][p.\ 94]{kroonen12} without
  being synonyms. However, within the 40-concept core vocabulary space covered by ASJP,
  such cross-concept cognate pairs are arguably very rare.} Table \ref{tab:4} illustrates
the resulting clustering for the concept ``person'' and a subset of the doculects.
 \begin{table}[h]
   \centering
   \begin{tabular}{l>{\tt } lc}
     \toprule
      doculect                                     & \normalfont{word}      & class label \\
     \midrule
     \rowcolor{lightgray}ALBANIAN                  & vet3      & 0           \\
     \rowcolor{lightgray}ALBANIAN\_TOSK            & vEt3      & 0           \\
     ARAGONESE                                     & ombre     & 1           \\
     \rowcolor{lightgray}ITALIAN\_GROSSETO\_TUSCAN & omo       & 2           \\
     \rowcolor{lightgray}ROMANIAN\_MEGLENO         & wom       & 2           \\
     \rowcolor{lightgray}VLACH                     & omu       & 2           \\
     ASTURIAN                                      & persona   & 3           \\
     BALEAR\_CATALAN                               & p3rson3   & 3           \\
     CATALAN                                       & p3rson3   & 3           \\
     FRIULIAN                                      & pErsoN    & 3           \\
     ITALIAN                                       & persona   & 3           \\
     SPANISH                                       & persona   & 3           \\
     VALENCIAN                                     & persone   & 3           \\
     \rowcolor{lightgray}CORSICAN                  & nimu      & 4           \\
     DALMATIAN                                     & om        & 5           \\
     EMILIANO\_CARPIGIANO                          & om        & 5           \\
     ROMANIAN\_2                                   & om        & 5           \\
     TURIA\_AROMANIAN                              & om        & 5           \\
     \rowcolor{lightgray}EMILIANO\_FERRARESE       & styan     & 6           \\
     \rowcolor{lightgray}LIGURIAN\_STELLA          & kristyaN  & 6           \\
     \rowcolor{lightgray}NEAPOLITAN\_CALABRESE     & kr3styan3 & 6           \\
     \rowcolor{lightgray}ROMAGNOL\_RAVENNATE       & sCan      & 6           \\
     \rowcolor{lightgray}ROMANSH\_GRISHUN          & k3rSTawn  & 6           \\
     \rowcolor{lightgray}ROMANSH\_SURMIRAN         & k3rstaN   & 6           \\
     GALICIAN                                      & ome       & 7           \\
     GASCON                                        & omi       & 7           \\
     \rowcolor{lightgray}PIEMONTESE\_VERCELLESE    & omaN      & 8           \\
     \rowcolor{lightgray}ROMANSH\_VALLADER         & uman      & 8           \\
     ALBANIAN\_GHEG                                & 5eri      & 9           \\
     SARDINIAN\_CAMPIDANESE                        & omini     & 9           \\
     SARDINIAN\_LOGUDARESE                         & omine     & 9           \\
     \bottomrule
   \end{tabular}
   \caption{Automatic cognate clustering for concept ``person''}
   \label{tab:4}
 \end{table}
A manual inspection reveals that the automatic classification does not completely coincide
with the cognate classification a human expert would assume. For instance, the descendants
of Latin \emph{homo} are split into classes 1, 2, 5, and 7. Also, Gheg Albanian
\texttt{5eri} and Sardinian \texttt{omini} have the same label but are not cognate.

Based on evaluations against manually assembled cognacy judgments for different but
similar data \citep{jaegerSofroniev16Konvens,jaegerListSofroniev17}, we can expect an
average F-score of 60\%--80\% for automatic cognate detection. This means that on average,
for each word, 60\%--80\% of its true cognates are assigned the same label, and 60\%--80\%
of the words carrying the same label are genuine cognates. 

\subsection{Phylogenetic inference}
\label{sec:phylogeneticInference}
\subsubsection{General remarks}

A \emph{phylogenetic tree} (or simply \emph{phylogeny}) is a similar data structure than
family trees according to the comparative method, but there are some subtle but important
differences between those concepts. Like a family tree, a phylogeny is a tree graph, i.e.,
an acyclic graph. If one node in the graph is identified as \emph{root}, the phylogeny is
\emph{rooted}; otherwise it is \emph{unrooted}. The branches (or edges) of a phylogeny
have non-negative \emph{branch lengths}. A phylogeny without branch length is called
\emph{topology}.

Nodes with a degree 1 (i.e., nodes which are the endpoint of exactly one branch) are
called \emph{leaves} or \emph{tips}. The are usually labeled with the names of observed
entities, such as documented languages. Nodes with a degree $>1$ are the \emph{internal
  nodes}. If the root (if present) has degree 2 and all other internal nodes have degree
3, the phylogeny is \emph{binary-branching.} Most algorithms for phylogenetic inference
produce binary-branching trees.

Like a linguistic family tree, a rooted phylogeny is a model of the historic process
leading to the observed diversity between the objects at the leaves. Time flows from the
root to the leaves. Internal nodes, represent unobserved historical objects, such as
ancient languages. Branching nodes represent \emph{diversification events}, i.e.\ the
splitting of a lineage into several daughter lineages.

The most important difference between family trees and phylogenies is the fact that the
latter have branch lengths. Depending on the context, these lengths may represent two
different quantities. They may capture the historic time (measured in years) between
diversification events, or they indicate the \emph{amount of change} along the branch,
measured for instance as the expected number of lexical replacements or the expected
number of sound changes. The two interpretations only coincide if the rate of change is
constant. This assumption is known to be invalid for language change (cf.\ e.g.\ the discussion
in \citealp{macmahonMacmahon2006}).

Another major difference, at least in practice, between family trees and phylogenies
concerns the type of justification that is expected for the stipulation of an internal
node. According to the comparative method, such a node is justified if and only if a
\emph{shared innovation} can be reconstructed for all daughter lineages of this
node.\footnote{``The only generally accepted criterion for subgrouping is \emph{shared
    innovation.}'' (\citealp{campbell1998}, p.\ 190, emphasis in original).} Consequently,
family trees obtained via the comparative method often contain multiply branching nodes
because the required evidence for further subgrouping is not available. Phylogenies, in
contradistinction, are mostly binary-branching, at least in practice. Partially this is a
matter of computational convenience since this reduces the search space. Also, algorithms
working recursively leaves-to-root can be formulated in a more efficient way if all
internal nodes are known to have at most two daughters. Furthermore, the degree of
justification of a certain topology is evaluated \emph{globally}, not for each internal
node individually. In the context of phylogenetic inference, it is therefore not required
to identify shared innovations for individual nodes.

There is a large variety of algorithms from computational biology to infer phylogenies
from observed data. The overarching theme of \emph{phylogenetic inference} is that a
phylogeny represents (or is part of) a mathematical model explaining the observed
variety. There are criteria quantifying how good an explanation a phylogeny provides for
observed data. Generally speaking, the goal is to find a phylogeny that provides an
optimal explanation for the observed data. The most commonly used algorithms are (in
ascending order of sophistication and computational costs) \emph{Neighbor Joining}
\citep{saitouNei87} and its variant \emph{BIONJ} \citep{bionj}, \emph{FastMe}
\citep{fastme}, \emph{Fitch-Margoliash} \citep{fitchMargoliash}, \emph{Maximum Parsimony}
\citep{mp}, \emph{Maximum Likelihood}\footnote{This method was developed incrementally;
  \citep{edwardsCavalliSforza64} is an early reference.} and \emph{Bayesian Phylogenetic
  Inference} (cf.\ \citealp{chenKuoLewis} for an overview).

The latter two approaches, \emph{Maximum Likelihood} and \emph{Bayesian Phylogenetic
  Inference} are based on a probabilistic model of language change. To apply them, a
language has to be represented as a \emph{character vector}. A \emph{character} is a
feature with a finite number of possible values, such as ``order of verb and object'',
``the first person plural pronoun contains a dental consonant'' or what have you. In most
applications, characters are binary, with ``0'' and ``1'' as possible values. In the
sequel, we will assume all characters are binary.

Diachronic change of a character value is modeled as a \emph{continuous time Markov
  process}. At each point in time a character can spontaneously switch to the other value
with a fixed probability density. A two-state process is characterized by two parameters,
$r$ and $s$, where $r$ is the \emph{rate of change} of $0\rightarrow 1$ (the probability
density of a switch to 1 if the current state is 0) and $s$ the rate of change for
$1\rightarrow 0$. For a given time interval of length $t$, the probability of being in
state $i$ at the start of the interval and in state $j$ at the end is then given by
$P(t)_{ij}$, where
\begin{eqnarray*}
  P(t) &=& 
           \frac{1}{r+s}\left(
           \begin{array}[c]{ll}
             s+re^{-(r+s)t}&r-re^{-(r+s)t}\\
             s-se^{-(r+s)t}&r+se^{-(r+s)t}
           \end{array}
           \right).
\end{eqnarray*}
The possibility of multiple switches occurring during the interval is factored in.

A probabilistic model for a given set of character vectors is a phylogenetic tree (with
the leaves indexed by the characters vectors) plus a mapping from edges to rates $(r,s)$
for each character and a probability distribution over character values at the root for
each character.

Suppose we know not only the character states at the leaves of the phylogeny but also at
all internal nodes. The likelihood of a given branch is then given by $P(t)_{ij}$, where
$i$ and $j$ are the states at the top and the bottom of the branch respectively, and $t$
is the length of the branch. The likelihood of the entire phylogeny for a given character
is then the product of all branch likelihoods, multiplied with the probability of the root
state. The total likelihood of the phylogeny is the product of its likelihoods for all
characters.

If only the character values for the leaves are known, the likelihood of the phylogeny
given the character vectors at the leaves is the sum of its likelihoods for all possible
state combinations at the internal nodes.

This general model is very parameter-rich since for each branch and each character, a pair
of rates have to be specified. There are various ways to reduce the degrees of
freedom. The simplest method is to assume that rates are constant across branches and
characters, and that the root probabilities of each character equal the equilibrium
probabilities of the Markov process: $(\nicefrac{s}{(r+s)}, \nicefrac{r}{(r+s)})$. More
sophisticated approaches assume that rates vary across characters and across branches
according to some parameter-poor probability distribution, and the expected likelihood of
the tree is obtained by integrating over this distribution. For a detailed mathematical
exposition, the interested reader is referred to the relevant literature from
computational biology, such as \citep{ewansGrant}.

A parameterized model, i.e., a phylogeny plus rate specifications for all characters and
branches, and root probabilities for each characters, assigns a certain likelihood to the
observed character vectors. \emph{Maximum Likelihood} (ML) inference searches for the
model that maximizes this likelihood given the observations. While the optimal numerical
parameters of a model, i.e., branch lengths, rates and root probabilities, can efficiently
be found by standard optimization techniques, finding the topology that gives rise to the
ML-model is computationally hard. Existing implementations use various heuristics to
search the tree space and find some local optimum, but there is no guarantee that the
globally optimal topology is found.\footnote{Among the best software packages currently
  available for ML phylogenetic inference are \emph{RAxML} \citep{raxml8} and
  \emph{IQ-Tree} \citep{iqtree}.}

\emph{Bayesian phylogenetic inference} requires some suitable prior probability
distributions over models (i.e., topologies, branch lengths, rates, possibly rate
variations across characters and rate variation across branches) and produces a sample of
the posterior distribution over models via a Markov Chain Monte Carlo
simulation.\footnote{Suitable software packages are, \emph{inter alia}, \emph{MrBayes}
  \citep{mrbayes3} and \emph{BEAST} \citep{beast2}.}

\subsubsection{Application to the case study}

For the case study, doculects were represented by two types of binary characters:
\begin{itemize}
\item \textbf{Inferred class label characters} (cf.\ Subsection \ref{sec:cc}). Each
  inferred class label is a character. A doculect has value 1 for such a character if and only if its
  word list contains a word carrying this label.\footnote{If a word list contains no entry
    for a certain concept, all characters pertaining to this concept are undefined for
    this concept. The same principle applies to the soundclass-concept characters. Leaves
    with undefined character values are disregarded when computing the likelihood of a
    phylogeny for that character.}
\item \textbf{Soundclass-concept characters}. There is one character for each pair $(s,c)$
  of a sound class $s$ and a concept $c$. A doculect has value 1 for that character if and
  only if its word list contains a word $w$ for $c$ that contains $s$ in its
  transcription.
\end{itemize}

Both types of characters carry a diachronic signal. For instance, the mutation
$0\rightarrow 1$ for class label 6/concept \emph{person} (cf. Table \ref{tab:4})
represents a lexical replacement of Latin ``homo'' or ``persona'' by descendants of Latin
``christianus'' in some Romance dialects \citep[p.\ 179]{meyerLuebke}. The mutation
$0\rightarrow 1$ for the soundclass-concept character \texttt{k}/\emph{person} represents
the same historical process. Soundclass-concept characters, however, also capture sound
shifts. For instance, the mutation $0\rightarrow 1$ for \texttt{b}/\emph{person} reflects
the epenthetic insertion of \texttt{b} in descendants of Latin ``homo'' in some Iberian
dialects.

We performed Bayesian phylogenetic inference on those characters. The inference was
carried out using the Software \emph{MrBayes} \citep{mrbayes3}. Separate rate models were
inferred for the two character types. Rate variation across characters was modeled by a
discretized Gamma distribution using 4 rate categories. We assumed no rate variation
across edges. Root probabilities were identified with equilibrium probabilities. An
ascertainment correction for missing all-0 characters was used.

We assumed rates to be constant across rates. This entails that the fitted branch lengths
reflect the expected amount of change (i.e., the expected number of mutations) along that
branch. 

In such a model, the likelihood of a phylogeny does not depend on the location of the
root (the assumed Markov process is \emph{time reversible}.) Therefore phylogenetic
inference provides no information about the location of the root. This motivates the
inclusion of the Albanian doculects. Those doculects were used as \emph{outgroup}, i.e.,
the root was placed on the branch separating the Albanian and the Romance doculects.

\begin{figure}[h]
  \centering
  \includegraphics[width=.8\linewidth]{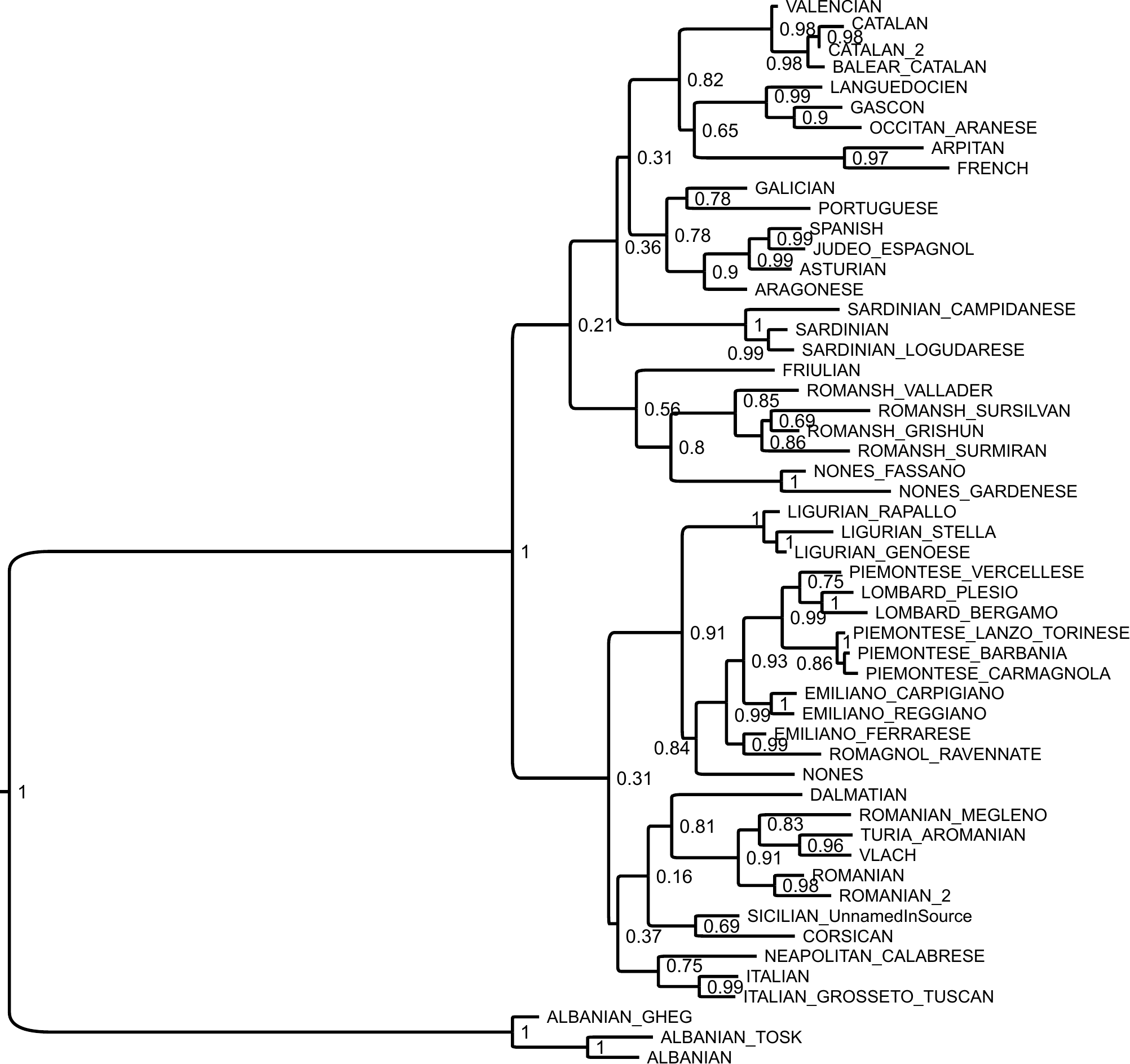}
  \caption{Representative phylogeny from the posterior distribution. Labels at the
    internal nodes indicate posterior probabilities}
  \label{fig:4}
\end{figure}

We obtained a sample of the posterior distribution containing 2,000 phylogenies. Figure
\ref{fig:4} displays a representative member of this sample (the \emph{maximum clade
  credibility} tree). The labels at the nodes indicate \emph{posterior probabilities} of
that node, i.e., the proportion of the phylogenies in the posterior sample having the same
sub-group.
These posterior probabilities are mostly rather low, indicating a large degree of
topological variation in the posterior sample. Some subgroups, such as Balkan Romance or
the Piemontese dialects, achieve high posterior probabilities though.

Notably, branch lengths carry information about the amount of change. According to the
phylogeny in Figure \ref{fig:4}, for instance, the Tuscan dialect of Italian
(ITALIAN\_GROSSETO\_TUSCAN) is predicted to be the most conservative Romance dialect
(since its distance to the latest common ancestors of all Romance dialects is shortest),
and French the most innovative one.

These results indicate that the data only contain a weak tree-like signal. This is
unsurprising since the Romance languages and dialects form a dialect continuum where
horizontal transfer of innovations is an important factor.

Phylogenetic trees, like traditional family trees, only model vertical descent, not
horizontal diffusion. They are therefore only an approximation of the historical
truth. But even though, they are useful as statistical models for further inference
steps. 

\subsection{Ancestral state reconstruction}
\label{sec:asr}

If a fully specified model is given, it is possible to estimate the probability
distributions over character states for each internal node.

Let $M=\langle\mathcal{T},\vec\theta\rangle$ be a model, i.e., a phylogeny $\mathcal{T}$
plus further parameters $\vec\theta$ (rates and root probabilities, possibly
specifications of rate variation). Let $i$ be a character and $n$ a node within
$\mathcal{T}$.

The parameters $\vec\theta$ specify a Markov process, including rates, for the branch
leading to $n$. Let $\langle\pi_0,\pi_1\rangle$ be the equilibrium probabilities of that
process. (If $n$ is the root, $\langle\pi_0,\pi_1\rangle$ are directly given by
$\vec\theta$.)

Let $M(n_i=x)$ be the same model as $M$, except that the value of character $i$ at node
$n$ is fixed to the value $x$. $\mathcal{L}(M)$ is the likelihood of model $M$ given the
observed character vectors for the leaves.

The probability distribution over values of character $i$ at node $n$, given $M$, is
determined by Bayes Rule:
\begin{eqnarray*}
  P(n_i=x|M) &\propto& \mathcal{L}(M(n_i=x))\times \pi_x
\end{eqnarray*}

\begin{figure}[h!]
  \centering
  \includegraphics[width=\linewidth]{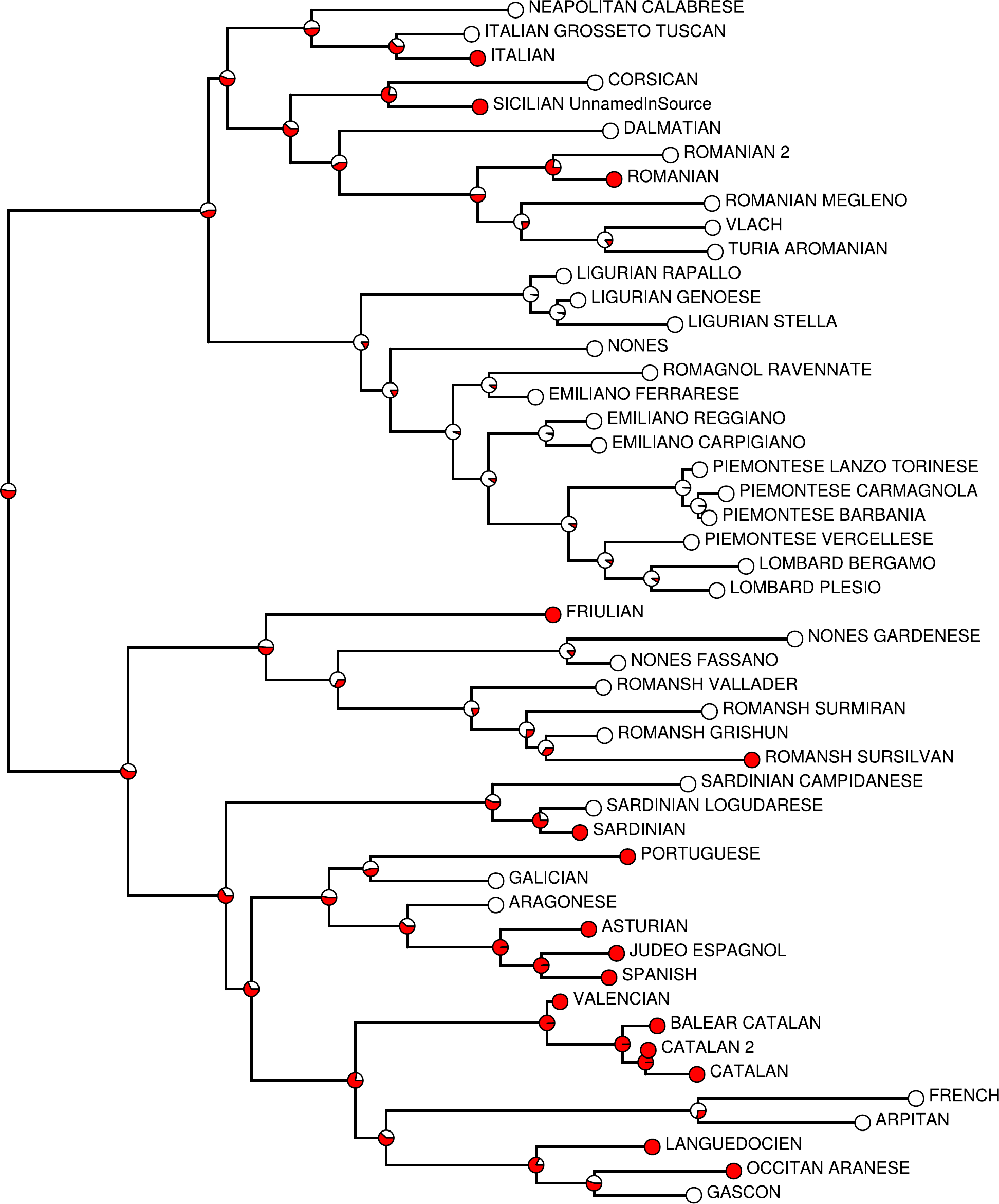}
  \caption{Ancestral state reconstruction for character \emph{person}:3}
  \label{fig:5}
\end{figure}
Figure \ref{fig:5} illustrates this principle with the Romance part of the tree from
Figure \ref{fig:4} and the character \emph{person}:3 (cf.\ Table \ref{tab:4}). The pie
charts at the nodes display the probability distribution for that node, where white
represents 0 and red 1.

This kind of computation was carried out for each class label character and each tree in
the posterior sample for the latest common ancestor of the Romance doculects. For each
concept, the class label for that concept with the highest average probability for value 1
at the root of the Romance subtree was inferred to represent the cognate class of the
Proto-Romance word for that concept.\footnote{See \citep{jaegerList17asr} for further
  elaboration and justification of this method of Ancestral State Reconstruction.} For the
concept \emph{person}, e.g., character \emph{person}:3 (representing the descendants of
Latin ``persona'') comes out as the best reconstruction.

\subsection{Multiple sequence alignment}

In the previous step, for the concept \emph{eye}, the class label 6 was reconstructed for
Proto-Romance. Its reflexes are given in Table \ref{tab:5}.
\begin{table}[h]
  \centering
  \begin{tabular}{l>{\tt } l}
    \toprule
    doculect                  & \normalfont{word} \\\midrule
    DALMATIAN                 & vaklo             \\
    ITALIAN                   & okkyo             \\
    ITALIAN\_GROSSETO\_TUSCAN & okyo              \\
    NEAPOLITAN\_CALABRESE     & wokyo             \\
    ROMANIAN\_2               & oky               \\
    ROMANIAN\_MEGLENO         & wokLu             \\
    SARDINIAN\_LOGUDARESE     & okru              \\
    SICILIAN\_UnnamedInSource & okiu              \\
    TURIA\_AROMANIAN          & okLu              \\
    VLACH                     & okklu             \\\bottomrule{}
  \end{tabular}
  \caption{Reflexes of class label \emph{eye}:6}
  \label{tab:5}
\end{table}

A \emph{multiple sequence alignment} (MSA) is a generalization of pairwise alignment to
more than two sequences. Ideally, all segments within a column are descendants of the
same sound in some common ancestor.

MSA, as applied to DNA or protein sequences, is a major research topic in
bioinformatics. The techniques developed in this field are \emph{mutatis mutandis} also
applicable to MSA of phonetic strings. In this Subsection one approach will briefly be
sketched. For a wider discussion and and proposals for related but different approaches,
see \citep{list14}.

Here we will follow the overall approach from \citep{tCoffee} and combine it with the
PMI-based method for pairwise alignment described in Subsection
\ref{sec:pairwise}. \citep{tCoffee} dub their approach \emph{T-Coffee} (``Tree-based
Consistency Objective Function For alignment Evaluation''), and we will use this name for
the method sketched here as well.

In a first step, all pairwise alignments between words from the list to be multiply
aligned are collected. For this purpose we use PMI pairwise alignment. Some examples would
be
\begin{center}\tt
  \begin{tabular}{llllll}
    okiu   & vaklo & okkyo & -okyo & o-ky-  & okru \\
    oky-   & wokLu & o-ky- & wokyo & okklu  & okiu \\
    $0.67$ & $0.2$ & $1.0$ & $1.0$ & $0.67$ & $0.75$ 
  \end{tabular}
\end{center}
The last row shows the \emph{score} of the alignment, i.e., the proportion of identical
matches (disregarding gaps).

In a second step, all \emph{indirect alignments} between a given word pair are collected,
which are obtained via relation composition with a third word. Some examples for indirect
alignments between \texttt{okiu} and \texttt{oky} would be:
\begin{center}\tt
  \begin{tabular}{lllll}
    okiu & -okiu & okiu & -okiu & oki-u \\
    okyo & wokyo & oky- & wokLu & okklu \\ 
    oky- & -oky- & oky- & -oky- & o-ky-
  \end{tabular}
\end{center}

The direct pairwise alignment matches the \emph{i} in \texttt{okiu} with the \texttt{y} in
\texttt{oky}. Most indirect alignments pair these two positions as well, but not all of
them. In the last columns, the \texttt{i} from \texttt{okiu} is related to the \texttt{k}
of \texttt{oky}, and the \texttt{y} from \texttt{oky} with a gap. For each pair of
positions in two strings, the relative frequency of them being indirectly aligned,
weighted by the score of the two pairwise alignments relating them, are summed. They form
the \emph{extended score} between those positions.

The optimal MSA for the entire group of words is the one were the sum of the pairwise
extended scores per column are maximized. Finding this global optimum is computationally
not feasible though, since the complexity of this task grows exponentially with the number
of sequences. \emph{Progressive alignment} \citep{progressiveAlignment} is a method to
obtain possibly sub-optimal but good MSAs in polynomial time. Using a \emph{guide tree}
with sequences at the leaves, MSAs are obtained recursively leaves-to-root. For each
internal node, the MSAs at the daughter nodes are combined via the Needleman-Wunsch
algorithm while respecting all partial alignments from the daughter nodes. 

For the words from Table \ref{tab:5}, this method produces the MSA in Table
\ref{tab:6}. The tree in Figure \ref{fig:4}, pruned to the doculects represented in the
word lists, was used as guide tree.
\begin{table}[h]
  \centering
  \begin{tabular}{l>{\tt } l}
    \toprule
    doculect                  & \normalfont{alignment} \\\midrule
    DALMATIAN                 & va-klo                 \\
    ITALIAN                   & -okkyo                 \\
    ITALIAN\_GROSSETO\_TUSCAN & -o-kyo                 \\
    NEAPOLITAN\_CALABRESE     & wo-kyo                 \\
    ROMANIAN\_2               & -o-ky-                 \\
    ROMANIAN\_MEGLENO         & wo-kLu                 \\
    SARDINIAN\_LOGUDARESE     & -o-kru                 \\
    SICILIAN\_UnnamedInSource & -o-kiu                 \\
    TURIA\_AROMANIAN          & -o-kLu                 \\
    VLACH                     & -okklu                 \\\bottomrule{}
  \end{tabular}
  \caption{Multiple Sequence Alignment for the word from Table \ref{tab:5}, using the tree
    from Figure \ref{fig:4} as guide tree}
  \label{tab:6}
\end{table}

Using this method MSAs were computed for each inferred class label that was inferred to be
present in Proto-Romance via Ancestral State Reconstruction.

\subsection{Proto-form reconstruction}

A final step toward the reconstruction of Proto-Romance forms, \emph{Ancestral State
  Reconstruction} is performed for the sound classes in each column, for each
MSA obtained in the previous step.

Consider the first column of the MSA in Table \ref{tab:5}. It contains three possible
states, \texttt{v}, \texttt{w}, and the gap symbols \texttt{-}. For each of these states,
a binary presence-absence character is constructed. For doculects which do not occur in
the MSA in question, this character is undefined.

The method for Ancestral State Reconstruction described in Subsection \ref{sec:asr} was
applied to these characters. For phylogeny in the posterior sample, the probabilities for
state 1 at the Proto-Romance node was computed for each character. For each column of an
MSA, the state with the highest average probability was considered as reconstructed.

The reconstructed proto-form for a given concept is then obtained by concatenating the
reconstructed states for the corresponding MSA and deleting all gap symbols. The results
are given in Table \ref{tab:7}.
\begin{table}[h!]
  \centering
  \begin{tabular}{>{\em} l>{\tt } l >{\tt }l}\toprule
    \normalfont{concept} & \normalfont{Latin}       & \normalfont{reconstruction} \\\midrule
    blood                & saNgw\textasciitilde{}is & saNg                        \\
    bone                 & os                       & os                          \\
    breast               & pektus, mama             & pet                         \\
    come                 & wenire                   & venir                       \\
    die                  & mori                     & murir                       \\
    dog                  & kanis                    & kan                         \\
    drink                & bibere                   & beb3r                       \\
    ear                  & auris                    & oreL3                       \\
    eye                  & okulus                   & okyu                        \\
    fire                 & iNnis                    & fok                         \\
    fish                 & piskis                   & peS                         \\
    full                 & plenus                   & plen                        \\
    hand                 & manus                    & man                         \\
    hear                 & audire                   & sentir                      \\
    horn                 & kornu                    & korn3                       \\
    I                    & ego                      & iy3                         \\
    knee                 & genu                     & Z3nuL                       \\
    leaf                 & foly\textasciitilde{}u*  & foLa                        \\
    liver                & yekur                    & figat                       \\
    louse                & pedikulus                & pidoko                      \\
    mountain             & mons                     & munta5a                     \\
    name                 & nomen                    & nom                         \\
    new                  & nowus                    & novo                        \\
    night                & noks                     & note                        \\
    nose                 & nasus                    & nas                         \\
    one                  & unus                     & unu                         \\
    path                 & viya                     & strada                      \\
    person               & persona, homo            & persona                     \\
    see                  & widere                   & veder                       \\
    skin                 & kutis                    & pel                         \\
    star                 & stela                    & stela                       \\
    stone                & lapis                    & pEtra                       \\
    sun                  & sol                      & sol                         \\
    tongue               & liNgw\textasciitilde{}E  & liNga                       \\
    tooth                & dens                     & dEnt                        \\
    tree                 & arbor                    & arbur                       \\
    two                  & duo                      & dos                         \\
    water                & akw\textasciitilde{}a    & akwa                        \\
    we                   & nos                      & nos                         \\
    you                  & tu                       & tu                          \\
  \end{tabular}
  \caption{Reconstructions for Proto-Romance}
  \label{tab:7}
\end{table}

\subsection{Evaluation}

\begin{figure}[h]
  \centering
  \includegraphics[width=.5\linewidth]{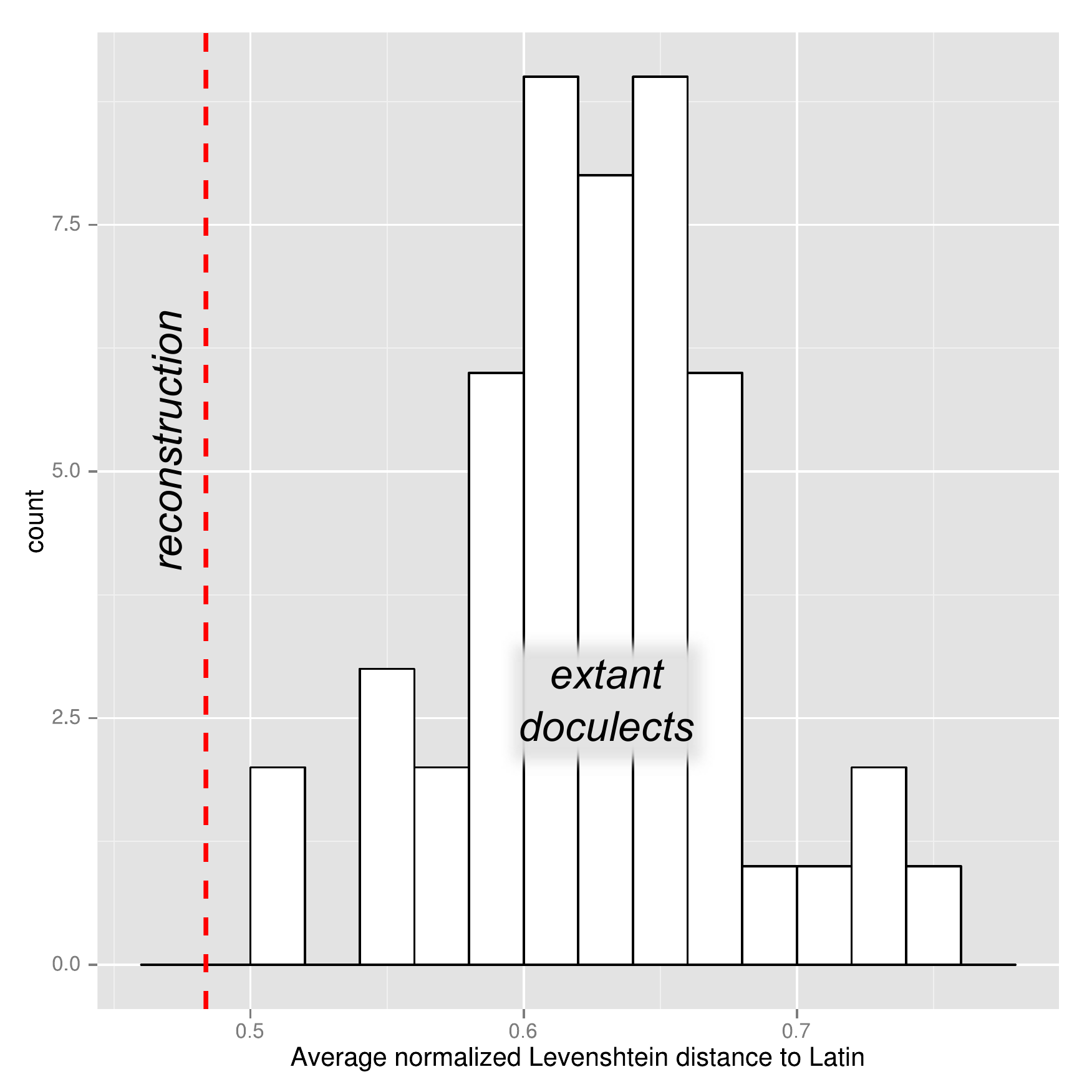}
  \caption{Average normalized Levenshtein distance to Latin words: reconstruction (dashed
    line) and extant Romance doculects (white bars)}
  \label{fig:6}
\end{figure}

To evaluate the quality of the automatic reconstructions, they were compared to the
corresponding elements of the Latin word list. For each reconstructed word, the normalized
Levenshtein distance (i.e., the Levenshtein distance divided by the length of the longer
string) to each Latin word (without diacritics) for that concept was computed. The
smallest such value counts as the score for that concept. The average score was
$0.484$. The extant Romance doculects have an average score of $0.627$. The most
conservative doculect, Sardinian, has a score of $0.502$, and the least conservative,
Arpitan, $0.742$. The evaluation results are depicted in Figure \ref{fig:6}.

These findings indicate that the automatic reconstruction does in fact capture a
historical signal. Manual inspection of the reconstructed word list reveals that to a
large degree, the discrepancies to Latin actually reflect language change between
Classical Latin and the latest common ancestor of the modern Romance doculects, namely
Vulgar Latin. To mention just a few points: (1) Modern Romance nouns are mostly derived
from the Latin accusative form \citep[p.\ 3]{hermanVulgarLatin}, while the word lists
contains the nominative form. For instance, the common ancestor forms for ``tooth'' and
``night'' are \emph{dentem} and \emph{noctem}. The reconstructed \texttt{t} in the
corresponding reconstructed forms are therefore historically correct. (2) Some Vulgar
Latin words are morphologically derived from their Classical Latin counterparts, such as
\emph{mons} $\rightarrow$ \emph{montanea} ``mountain'' \citep[p.\ 464]{meyerLuebke} or
\emph{genus} $\rightarrow$ \emph{genukulum} ``knee'' \citep[p.\ 319]{meyerLuebke}. This is
likewise partially reflected in the reconstructions. (3) For some concepts, lexical
replacement by non-cognate words took place between Classical and Vulgar Latin, such as
\emph{via} $\rightarrow$ \emph{strata} ``path'',\footnote{Latin makes a semantic
  distinction between \emph{via} for unpaved and \emph{strata} for paved roads; cf.\
  \citep[p.\ 685]{meyerLuebke}} \emph{ignis} $\rightarrow$ \emph{focus} ``fire''
\citep[p.\ 293]{meyerLuebke}, or \emph{iecur} $\rightarrow$ \emph{ficatum} ``liver''
\citep[p.\ 106]{hermanVulgarLatin}. Again, this is reflected in the reconstruction.

On the negative side, the reconstructions occasionally reflect sound changes that only
took place in the Western Romania, such as the voicing of plosives between vowels
\citep[p.\ 46]{hermanVulgarLatin}.

Let us conclude this section with some reflections on how the reconstructions were
obtained and how this relates to the comparative method.

A major difference to the traditional approach is the stochastic nature of the workflow
sketched here. Both phylogenetic inference and ancestral state reconstruction is based on
probabilities rather than categorical decisions. The results shown in Table \ref{tab:7}
propose a unique reconstruction for each concept, but it would be a minor modification of
the workflow only to derive a probability distribution over reconstructions instead. This
probabilistic approach is arguably an advantage since it allows to utilize uncertain and
inconclusive information while taking this uncertainty properly into account.

Another major difference concerns the multiple independence assumptions implicit in the
probabilistic model sketched in Subsection \ref{sec:phylogeneticInference}. The likelihood
of a phylogeny is the product of its likelihoods for the individual characters. This
amounts to the assumptions that the characters are mutually stochastically independent.

For the characters used here (and generally in computational phylogenetics as applied to
historical linguistics) are mutually dependent in manifold ways though. For instance, the
loss of a cognate class makes it more likely that the affected lineage will acquire
another cognate class for the same semantic slot and vice versa. 

This problem is even more severe for phonetic change. Since the work of the Neogrammarians
in the 19th century, it is recognized that many sound changes are \emph{regular}, i.e.,
they apply to all instances of a certain sound (given contextual conditions) throughout
the lexicon. Furthermore, both regular and irregular sound changes are usually dependent
on their syntagmatic phonetic context, and sometimes on the paradigmatic context within
inflectional paradigms as well. \citep{bouchardetal13} and \citep{hruschkaetal15} propose
more sophisticated probabilistic models of language change than the one used here to take
these dependencies into account.\footnote{So far these model have only been tested only on
  one language family each (Austronesian and Turkic respectively), and the algorithmic
  tools have not been released.}

Last but not least, the treatment of borrowing (and language contact in general) are an
unsolved problem for computational historical linguistics. Automatic cognate clustering
does not distinguish between genuine cognates (related via unbroken chains of vertical
descent) and (descendants of) loanwords. This introduces a potential bias for phylogenetic
inference and ancestral state reconstruction, since borrowed items might be misconstrued
as shared retentions.

\section{Conclusion}

This article give a brief sketch of the state of the art in computational historical
linguistics, a relatively young subfield at the interface between historical linguistics,
computational linguistics and computational biology. The case study discussed in the
previous section serves to illustrate some of the major research topics in this domain:
identification of genetic relationships between languages, phylogenetic inference,
automatic cognate detection and ancestral state recognition. These concern the core issues
of the field; the information obtained by these methods are suitable to address questions
of greater generality, pertaining to general patterns of language change as well as the
relationship between the linguistic and non-linguistic history of specific human
populations.

\section*{Appendix}

All code and data used and produced when conduction the case study in Section
\ref{sec:protoRomance} are available for download and inspection from
\url{https://github.com/gerhardJaeger/protoRomance}.

\begingroup
\let\itshape\upshape
\let\emshape\upshape
\setlength{\bibsep}{0pt}

\endgroup

\begin{thebibliography}{}

\bibitem [\protect \citeauthoryear {%
Anthony%
}{%
Anthony%
}{%
{\protect \APACyear {2010}}%
}]{%
anthony2010}
\APACinsertmetastar {%
anthony2010}%
\begin{APACrefauthors}%
Anthony, D\BPBI W.%
\end{APACrefauthors}%
\unskip\
\newblock
\APACrefYear{2010}.
\newblock
\APACrefbtitle {The horse, the wheel, and language: how Bronze-Age riders from
  the {Eurasian} steppes shaped the modern world} {The horse, the wheel, and
  language: how bronze-age riders from the {Eurasian} steppes shaped the modern
  world}.
\newblock
\APACaddressPublisher{Princeton}{Princeton University Press}.
\PrintBackRefs{\CurrentBib}

\bibitem [\protect \citeauthoryear {%
Atkinson%
, Meade%
, Venditti%
, Greenhill%
\BCBL {}\ \BBA {} Pagel%
}{%
Atkinson%
\ \protect \BOthers {.}}{%
{\protect \APACyear {2008}}%
}]{%
atkinsonetal2008}
\APACinsertmetastar {%
atkinsonetal2008}%
\begin{APACrefauthors}%
Atkinson, Q\BPBI D.%
, Meade, A.%
, Venditti, C.%
, Greenhill, S\BPBI J.%
\BCBL {}\ \BBA {} Pagel, M.%
\end{APACrefauthors}%
\unskip\
\newblock
\APACrefYearMonthDay{2008}{}{}.
\newblock
{\BBOQ}\APACrefatitle {Languages evolve in punctuational bursts} {Languages
  evolve in punctuational bursts}.{\BBCQ}
\newblock
\APACjournalVolNumPages{Science}{319}{5863}{588--588}.
\PrintBackRefs{\CurrentBib}

\bibitem [\protect \citeauthoryear {%
Baxter%
\ \BBA {} {Manaster Ramer}%
}{%
Baxter%
\ \BBA {} {Manaster Ramer}%
}{%
{\protect \APACyear {2000}}%
}]{%
baxterManasterRamer2000}
\APACinsertmetastar {%
baxterManasterRamer2000}%
\begin{APACrefauthors}%
Baxter, W\BPBI H.%
\BCBT {}\ \BBA {} {Manaster Ramer}, A.%
\end{APACrefauthors}%
\unskip\
\newblock
\APACrefYearMonthDay{2000}{}{}.
\newblock
{\BBOQ}\APACrefatitle {{Beyond lumping and splitting. Probabilistic issues in
  historical linguistics}} {{Beyond lumping and splitting. Probabilistic issues
  in historical linguistics}}.{\BBCQ}
\newblock
\BIn{} C.~Renfrew, A.~McMahon\BCBL {}\ \BBA {} L.~Trask\ (\BEDS),
  \APACrefbtitle {Time depth in historical linguistics} {Time depth in
  historical linguistics}\ (\BVOL~1, \BPGS\ 167--188).
\newblock
\APACaddressPublisher{Cambridge}{McDonald Institute for Archaeological
  Research}.
\PrintBackRefs{\CurrentBib}

\bibitem [\protect \citeauthoryear {%
Bergsma%
\ \BBA {} Kondrak%
}{%
Bergsma%
\ \BBA {} Kondrak%
}{%
{\protect \APACyear {2007}}%
}]{%
Bergsma2007}
\APACinsertmetastar {%
Bergsma2007}%
\begin{APACrefauthors}%
Bergsma, S.%
\BCBT {}\ \BBA {} Kondrak, G.%
\end{APACrefauthors}%
\unskip\
\newblock
\APACrefYearMonthDay{2007}{}{}.
\newblock
{\BBOQ}\APACrefatitle {Multilingual cognate identification using integer linear
  programming} {Multilingual cognate identification using integer linear
  programming}.{\BBCQ}
\newblock
\BIn{} \APACrefbtitle {Proceedings of the RANLP Workshop} {Proceedings of the
  ranlp workshop}\ (\BPG~656-663).
\PrintBackRefs{\CurrentBib}

\bibitem [\protect \citeauthoryear {%
Bouchard-C{\^o}t{\'e}%
, Hall%
, Griffiths%
\BCBL {}\ \BBA {} Klein%
}{%
Bouchard-C{\^o}t{\'e}%
\ \protect \BOthers {.}}{%
{\protect \APACyear {2013}}%
}]{%
bouchardetal13}
\APACinsertmetastar {%
bouchardetal13}%
\begin{APACrefauthors}%
Bouchard-C{\^o}t{\'e}, A.%
, Hall, D.%
, Griffiths, T\BPBI L.%
\BCBL {}\ \BBA {} Klein, D.%
\end{APACrefauthors}%
\unskip\
\newblock
\APACrefYearMonthDay{2013}{}{}.
\newblock
{\BBOQ}\APACrefatitle {Automated reconstruction of ancient languages using
  probabilistic models of sound change} {Automated reconstruction of ancient
  languages using probabilistic models of sound change}.{\BBCQ}
\newblock
\APACjournalVolNumPages{Proceedings of the National Academy of
  Sciences}{36}{2}{141--150}.
\PrintBackRefs{\CurrentBib}

\bibitem [\protect \citeauthoryear {%
Bouckaert%
\ \protect \BOthers {.}}{%
Bouckaert%
\ \protect \BOthers {.}}{%
{\protect \APACyear {2014}}%
}]{%
beast2}
\APACinsertmetastar {%
beast2}%
\begin{APACrefauthors}%
Bouckaert, R.%
, Heled, J.%
, K{\"u}hnert, D.%
, Vaughan, T.%
, Wu, C\BHBI H.%
, Xie, D.%
\BDBL {}Drummond, A\BPBI J.%
\end{APACrefauthors}%
\unskip\
\newblock
\APACrefYearMonthDay{2014}{}{}.
\newblock
{\BBOQ}\APACrefatitle {{BEAST} 2: a software platform for {Bayesian}
  evolutionary analysis} {{BEAST} 2: a software platform for {Bayesian}
  evolutionary analysis}.{\BBCQ}
\newblock
\APACjournalVolNumPages{PLoS Computational Biology}{10}{4}{e1003537}.
\PrintBackRefs{\CurrentBib}

\bibitem [\protect \citeauthoryear {%
Bouckaert%
\ \protect \BOthers {.}}{%
Bouckaert%
\ \protect \BOthers {.}}{%
{\protect \APACyear {2012}}%
}]{%
bouckaertetal12}
\APACinsertmetastar {%
bouckaertetal12}%
\begin{APACrefauthors}%
Bouckaert, R.%
, Lemey, P.%
, Dunn, M.%
, Greenhill, S\BPBI J.%
, Alekseyenko, A\BPBI V.%
, Drummond, A\BPBI J.%
\BDBL {}Atkinson, Q\BPBI D.%
\end{APACrefauthors}%
\unskip\
\newblock
\APACrefYearMonthDay{2012}{}{}.
\newblock
{\BBOQ}\APACrefatitle {Mapping the origins and expansion of the
  {Indo}-{European} language family} {Mapping the origins and expansion of the
  {Indo}-{European} language family}.{\BBCQ}
\newblock
\APACjournalVolNumPages{Science}{337}{6097}{957--960}.
\PrintBackRefs{\CurrentBib}

\bibitem [\protect \citeauthoryear {%
Brown%
, Holman%
\BCBL {}\ \BBA {} Wichmann%
}{%
Brown%
\ \protect \BOthers {.}}{%
{\protect \APACyear {2013}}%
}]{%
brownetal13}
\APACinsertmetastar {%
brownetal13}%
\begin{APACrefauthors}%
Brown, C\BPBI H.%
, Holman, E.%
\BCBL {}\ \BBA {} Wichmann, S.%
\end{APACrefauthors}%
\unskip\
\newblock
\APACrefYearMonthDay{2013}{}{}.
\newblock
{\BBOQ}\APACrefatitle {Sound correspondences in the world's languages} {Sound
  correspondences in the world's languages}.{\BBCQ}
\newblock
\APACjournalVolNumPages{Language}{89}{1}{4--29}.
\PrintBackRefs{\CurrentBib}

\bibitem [\protect \citeauthoryear {%
Campbell%
}{%
Campbell%
}{%
{\protect \APACyear {1998}}%
}]{%
campbell1998}
\APACinsertmetastar {%
campbell1998}%
\begin{APACrefauthors}%
Campbell, L.%
\end{APACrefauthors}%
\unskip\
\newblock
\APACrefYear{1998}.
\newblock
\APACrefbtitle {Historical Linguistics. An Introduction} {Historical
  linguistics. an introduction}.
\newblock
\APACaddressPublisher{Edinburgh}{Edinburgh University Press}.
\PrintBackRefs{\CurrentBib}

\bibitem [\protect \citeauthoryear {%
Chen%
, Kuo%
\BCBL {}\ \BBA {} Lewis%
}{%
Chen%
\ \protect \BOthers {.}}{%
{\protect \APACyear {2014}}%
}]{%
chenKuoLewis}
\APACinsertmetastar {%
chenKuoLewis}%
\begin{APACrefauthors}%
Chen, M\BHBI H.%
, Kuo, L.%
\BCBL {}\ \BBA {} Lewis, P\BPBI O.%
\end{APACrefauthors}%
\unskip\
\newblock
\APACrefYear{2014}.
\newblock
\APACrefbtitle {Bayesian Phylogenetics. Methods, Algorithms and Applications}
  {Bayesian phylogenetics. methods, algorithms and applications}.
\newblock
\APACaddressPublisher{Abingdon}{CRC Press}.
\PrintBackRefs{\CurrentBib}

\bibitem [\protect \citeauthoryear {%
Covington%
}{%
Covington%
}{%
{\protect \APACyear {1996}}%
}]{%
covington96}
\APACinsertmetastar {%
covington96}%
\begin{APACrefauthors}%
Covington, M\BPBI A.%
\end{APACrefauthors}%
\unskip\
\newblock
\APACrefYearMonthDay{1996}{}{}.
\newblock
{\BBOQ}\APACrefatitle {An algorithm to align words for historical comparison}
  {An algorithm to align words for historical comparison}.{\BBCQ}
\newblock
\APACjournalVolNumPages{Computational linguistics}{22}{4}{481--496}.
\PrintBackRefs{\CurrentBib}

\bibitem [\protect \citeauthoryear {%
Csardi%
\ \BBA {} Nepusz%
}{%
Csardi%
\ \BBA {} Nepusz%
}{%
{\protect \APACyear {2006}}%
}]{%
igraph}
\APACinsertmetastar {%
igraph}%
\begin{APACrefauthors}%
Csardi, G.%
\BCBT {}\ \BBA {} Nepusz, T.%
\end{APACrefauthors}%
\unskip\
\newblock
\APACrefYearMonthDay{2006}{}{}.
\newblock
{\BBOQ}\APACrefatitle {The igraph software package for complex network
  research} {The igraph software package for complex network research}.{\BBCQ}
\newblock
\APACjournalVolNumPages{InterJournal, Complex Systems}{1695}{5}{1--9}.
\PrintBackRefs{\CurrentBib}

\bibitem [\protect \citeauthoryear {%
Desper%
\ \BBA {} Gascuel%
}{%
Desper%
\ \BBA {} Gascuel%
}{%
{\protect \APACyear {2002}}%
}]{%
fastme}
\APACinsertmetastar {%
fastme}%
\begin{APACrefauthors}%
Desper, R.%
\BCBT {}\ \BBA {} Gascuel, O.%
\end{APACrefauthors}%
\unskip\
\newblock
\APACrefYearMonthDay{2002}{}{}.
\newblock
{\BBOQ}\APACrefatitle {Fast and accurate phylogeny reconstruction algorithms
  based on the minimum-evolution principle} {Fast and accurate phylogeny
  reconstruction algorithms based on the minimum-evolution principle}.{\BBCQ}
\newblock
\APACjournalVolNumPages{Journal of computational biology}{9}{5}{687--705}.
\PrintBackRefs{\CurrentBib}

\bibitem [\protect \citeauthoryear {%
Dolgopolsky%
}{%
Dolgopolsky%
}{%
{\protect \APACyear {1986}}%
}]{%
dolgopolsky86}
\APACinsertmetastar {%
dolgopolsky86}%
\begin{APACrefauthors}%
Dolgopolsky, A\BPBI B.%
\end{APACrefauthors}%
\unskip\
\newblock
\APACrefYearMonthDay{1986}{}{}.
\newblock
{\BBOQ}\APACrefatitle {A probabilistic hypothesis concerning the oldest
  relationships among the language families of {Northern} {Eurasia}} {A
  probabilistic hypothesis concerning the oldest relationships among the
  language families of {Northern} {Eurasia}}.{\BBCQ}
\newblock
\BIn{} V\BPBI V.~Shevoroshkin\ (\BED), \APACrefbtitle {Typology, Relationship
  and Time: A collection of papers on language change and relationship by
  Soviet linguists} {Typology, relationship and time: A collection of papers on
  language change and relationship by soviet linguists}\ (\BPGS\ 27--50).
\newblock
\APACaddressPublisher{Ann Arbor}{Karoma Publisher}.
\PrintBackRefs{\CurrentBib}

\bibitem [\protect \citeauthoryear {%
Dunn%
, Greenhill%
, Levinson%
\BCBL {}\ \BBA {} Gray%
}{%
Dunn%
\ \protect \BOthers {.}}{%
{\protect \APACyear {2011}}%
}]{%
dunnetal11}
\APACinsertmetastar {%
dunnetal11}%
\begin{APACrefauthors}%
Dunn, M.%
, Greenhill, S\BPBI J.%
, Levinson, S.%
\BCBL {}\ \BBA {} Gray, R\BPBI D.%
\end{APACrefauthors}%
\unskip\
\newblock
\APACrefYearMonthDay{2011}{}{}.
\newblock
{\BBOQ}\APACrefatitle {Evolved structure of language shows lineage-specific
  trends in word-order universals} {Evolved structure of language shows
  lineage-specific trends in word-order universals}.{\BBCQ}
\newblock
\APACjournalVolNumPages{Nature}{473}{7345}{79--82}.
\PrintBackRefs{\CurrentBib}

\bibitem [\protect \citeauthoryear {%
Durbin%
, Eddy%
, Krogh%
\BCBL {}\ \BBA {} Mitchison%
}{%
Durbin%
\ \protect \BOthers {.}}{%
{\protect \APACyear {1989}}%
}]{%
durbinetal98}
\APACinsertmetastar {%
durbinetal98}%
\begin{APACrefauthors}%
Durbin, R.%
, Eddy, S\BPBI R.%
, Krogh, A.%
\BCBL {}\ \BBA {} Mitchison, G.%
\end{APACrefauthors}%
\unskip\
\newblock
\APACrefYear{1989}.
\newblock
\APACrefbtitle {Biological Sequence Analysis} {Biological sequence analysis}.
\newblock
\APACaddressPublisher{Cambridge, UK}{Cambridge University Press}.
\PrintBackRefs{\CurrentBib}

\bibitem [\protect \citeauthoryear {%
Dyen%
, Kruskal%
\BCBL {}\ \BBA {} Black%
}{%
Dyen%
\ \protect \BOthers {.}}{%
{\protect \APACyear {1992}}%
}]{%
kruskalDyenBlack92}
\APACinsertmetastar {%
kruskalDyenBlack92}%
\begin{APACrefauthors}%
Dyen, I.%
, Kruskal, J\BPBI B.%
\BCBL {}\ \BBA {} Black, P.%
\end{APACrefauthors}%
\unskip\
\newblock
\APACrefYearMonthDay{1992}{}{}.
\newblock
{\BBOQ}\APACrefatitle {An {Indoeuropean} Classification: A Lexicostatistical
  Experiment} {An {Indoeuropean} classification: A lexicostatistical
  experiment}.{\BBCQ}
\newblock
\APACjournalVolNumPages{Transactions of the American Philosophical
  Society}{82}{5}{1--132}.
\PrintBackRefs{\CurrentBib}

\bibitem [\protect \citeauthoryear {%
Edwards%
\ \BBA {} Cavalli-Sforza%
}{%
Edwards%
\ \BBA {} Cavalli-Sforza%
}{%
{\protect \APACyear {1964}}%
}]{%
edwardsCavalliSforza64}
\APACinsertmetastar {%
edwardsCavalliSforza64}%
\begin{APACrefauthors}%
Edwards, A\BPBI W\BPBI F.%
\BCBT {}\ \BBA {} Cavalli-Sforza, L\BPBI L.%
\end{APACrefauthors}%
\unskip\
\newblock
\APACrefYearMonthDay{1964}{}{}.
\newblock
{\BBOQ}\APACrefatitle {Reconstruction of evolutionary trees} {Reconstruction of
  evolutionary trees}.{\BBCQ}
\newblock
\BIn{} V\BPBI H.~Heywood\ \BBA {} J\BPBI R.~McNeill\ (\BEDS), \APACrefbtitle
  {Phenetic and Phylogenetic Classification} {Phenetic and phylogenetic
  classification}\ (\BPGS\ 67--76).
\newblock
\APACaddressPublisher{London}{Systematics Association Publisher}.
\PrintBackRefs{\CurrentBib}

\bibitem [\protect \citeauthoryear {%
Embleton%
}{%
Embleton%
}{%
{\protect \APACyear {1986}}%
}]{%
embleton1986}
\APACinsertmetastar {%
embleton1986}%
\begin{APACrefauthors}%
Embleton, S\BPBI M.%
\end{APACrefauthors}%
\unskip\
\newblock
\APACrefYear{1986}.
\newblock
\APACrefbtitle {Statistics in historical linguistics} {Statistics in historical
  linguistics}.
\newblock
\APACaddressPublisher{Bochum}{Brockmeyer}.
\PrintBackRefs{\CurrentBib}

\bibitem [\protect \citeauthoryear {%
Ewens%
\ \BBA {} Grant%
}{%
Ewens%
\ \BBA {} Grant%
}{%
{\protect \APACyear {2005}}%
}]{%
ewansGrant}
\APACinsertmetastar {%
ewansGrant}%
\begin{APACrefauthors}%
Ewens, W.%
\BCBT {}\ \BBA {} Grant, G.%
\end{APACrefauthors}%
\unskip\
\newblock
\APACrefYear{2005}.
\newblock
\APACrefbtitle {Statistical Methods in Bioinformatics: An Introduction}
  {Statistical methods in bioinformatics: An introduction}.
\newblock
\APACaddressPublisher{New York}{Springer}.
\PrintBackRefs{\CurrentBib}

\bibitem [\protect \citeauthoryear {%
Fitch%
}{%
Fitch%
}{%
{\protect \APACyear {1971}}%
}]{%
mp}
\APACinsertmetastar {%
mp}%
\begin{APACrefauthors}%
Fitch, W\BPBI M.%
\end{APACrefauthors}%
\unskip\
\newblock
\APACrefYearMonthDay{1971}{}{}.
\newblock
{\BBOQ}\APACrefatitle {Toward defining the course of evolution: minimum change
  for a specific tree topology} {Toward defining the course of evolution:
  minimum change for a specific tree topology}.{\BBCQ}
\newblock
\APACjournalVolNumPages{Systematic Zoology}{20}{4}{406--416}.
\PrintBackRefs{\CurrentBib}

\bibitem [\protect \citeauthoryear {%
Fitch%
\ \BBA {} Margoliash%
}{%
Fitch%
\ \BBA {} Margoliash%
}{%
{\protect \APACyear {1967}}%
}]{%
fitchMargoliash}
\APACinsertmetastar {%
fitchMargoliash}%
\begin{APACrefauthors}%
Fitch, W\BPBI M.%
\BCBT {}\ \BBA {} Margoliash, E.%
\end{APACrefauthors}%
\unskip\
\newblock
\APACrefYearMonthDay{1967}{}{}.
\newblock
{\BBOQ}\APACrefatitle {Construction of Phylogenetic Trees} {Construction of
  phylogenetic trees}.{\BBCQ}
\newblock
\APACjournalVolNumPages{Science}{155}{3760}{279-284}.
\PrintBackRefs{\CurrentBib}

\bibitem [\protect \citeauthoryear {%
Gascuel%
}{%
Gascuel%
}{%
{\protect \APACyear {1997}}%
}]{%
bionj}
\APACinsertmetastar {%
bionj}%
\begin{APACrefauthors}%
Gascuel, O.%
\end{APACrefauthors}%
\unskip\
\newblock
\APACrefYearMonthDay{1997}{}{}.
\newblock
{\BBOQ}\APACrefatitle {{BIONJ}: An improved version of the {NJ} algorithm based
  on a simple model of sequence data} {{BIONJ}: An improved version of the {NJ}
  algorithm based on a simple model of sequence data}.{\BBCQ}
\newblock
\APACjournalVolNumPages{Molecular Biology and Evolution}{14}{7}{685--695}.
\PrintBackRefs{\CurrentBib}

\bibitem [\protect \citeauthoryear {%
Gray%
\ \BBA {} Atkinson%
}{%
Gray%
\ \BBA {} Atkinson%
}{%
{\protect \APACyear {2003}}%
}]{%
grayAtkinson03}
\APACinsertmetastar {%
grayAtkinson03}%
\begin{APACrefauthors}%
Gray, R\BPBI D.%
\BCBT {}\ \BBA {} Atkinson, Q\BPBI D.%
\end{APACrefauthors}%
\unskip\
\newblock
\APACrefYearMonthDay{2003}{}{}.
\newblock
{\BBOQ}\APACrefatitle {Language-tree divergence times support the {Anatolian}
  theory of {Indo}-{European} origin} {Language-tree divergence times support
  the {Anatolian} theory of {Indo}-{European} origin}.{\BBCQ}
\newblock
\APACjournalVolNumPages{Nature}{426}{27}{435--439}.
\PrintBackRefs{\CurrentBib}

\bibitem [\protect \citeauthoryear {%
Gray%
, Drummond%
\BCBL {}\ \BBA {} Greenhill%
}{%
Gray%
\ \protect \BOthers {.}}{%
{\protect \APACyear {2009}}%
}]{%
grayDrummondGreenhill09}
\APACinsertmetastar {%
grayDrummondGreenhill09}%
\begin{APACrefauthors}%
Gray, R\BPBI D.%
, Drummond, A\BPBI J.%
\BCBL {}\ \BBA {} Greenhill, S\BPBI J.%
\end{APACrefauthors}%
\unskip\
\newblock
\APACrefYearMonthDay{2009}{}{}.
\newblock
{\BBOQ}\APACrefatitle {Language phylogenies reveal expansion pulses and pauses
  in {Pacific} settlement} {Language phylogenies reveal expansion pulses and
  pauses in {Pacific} settlement}.{\BBCQ}
\newblock
\APACjournalVolNumPages{Science}{323}{5913}{479--483}.
\PrintBackRefs{\CurrentBib}

\bibitem [\protect \citeauthoryear {%
Gray%
\ \BBA {} Jordan%
}{%
Gray%
\ \BBA {} Jordan%
}{%
{\protect \APACyear {2000}}%
}]{%
grayJordan2000}
\APACinsertmetastar {%
grayJordan2000}%
\begin{APACrefauthors}%
Gray, R\BPBI D.%
\BCBT {}\ \BBA {} Jordan, F\BPBI M.%
\end{APACrefauthors}%
\unskip\
\newblock
\APACrefYearMonthDay{2000}{}{}.
\newblock
{\BBOQ}\APACrefatitle {Language trees support the express-train sequence of
  {Austronesian} expansion} {Language trees support the express-train sequence
  of {Austronesian} expansion}.{\BBCQ}
\newblock
\APACjournalVolNumPages{Nature}{405}{6790}{1052--1055}.
\PrintBackRefs{\CurrentBib}

\bibitem [\protect \citeauthoryear {%
Greenhill%
, Blust%
\BCBL {}\ \BBA {} Gray%
}{%
Greenhill%
\ \protect \BOthers {.}}{%
{\protect \APACyear {2008}}%
}]{%
abvd}
\APACinsertmetastar {%
abvd}%
\begin{APACrefauthors}%
Greenhill, S\BPBI J.%
, Blust, R.%
\BCBL {}\ \BBA {} Gray, R\BPBI D.%
\end{APACrefauthors}%
\unskip\
\newblock
\APACrefYearMonthDay{2008}{}{}.
\newblock
{\BBOQ}\APACrefatitle {The {Austronesian} {Basic} {Vocabulary} {Database}: From
  Bioinformatics to Lexomics} {The {Austronesian} {Basic} {Vocabulary}
  {Database}: From bioinformatics to lexomics}.{\BBCQ}
\newblock
\APACjournalVolNumPages{Evolutionary Bioinformatics}{4}{}{271--283}.
\PrintBackRefs{\CurrentBib}

\bibitem [\protect \citeauthoryear {%
Haak%
\ \protect \BOthers {.}}{%
Haak%
\ \protect \BOthers {.}}{%
{\protect \APACyear {2015}}%
}]{%
haaketal2015}
\APACinsertmetastar {%
haaketal2015}%
\begin{APACrefauthors}%
Haak, W.%
, Lazaridis, I.%
, Patterson, N.%
, Rohland, N.%
, Mallick, S.%
, Llamas, B.%
\BDBL {}Reich, D.%
\end{APACrefauthors}%
\unskip\
\newblock
\APACrefYearMonthDay{2015}{}{}.
\newblock
{\BBOQ}\APACrefatitle {Massive migration from the steppe was a source for
  {Indo-European} languages in {Europe}} {Massive migration from the steppe was
  a source for {Indo-European} languages in {Europe}}.{\BBCQ}
\newblock
\APACjournalVolNumPages{Nature}{522}{7555}{207--211}.
\PrintBackRefs{\CurrentBib}

\bibitem [\protect \citeauthoryear {%
Hall%
\ \BBA {} Klein%
}{%
Hall%
\ \BBA {} Klein%
}{%
{\protect \APACyear {2010}}%
}]{%
hallKlein2010}
\APACinsertmetastar {%
hallKlein2010}%
\begin{APACrefauthors}%
Hall, D.%
\BCBT {}\ \BBA {} Klein, D.%
\end{APACrefauthors}%
\unskip\
\newblock
\APACrefYearMonthDay{2010}{}{}.
\newblock
{\BBOQ}\APACrefatitle {Finding cognate groups using phylogenies} {Finding
  cognate groups using phylogenies}.{\BBCQ}
\newblock
\BIn{} \APACrefbtitle {Proceedings of the 48th Annual Meeting of the
  {Association} for {Computational} {Linguistics}} {Proceedings of the 48th
  annual meeting of the {Association} for {Computational} {Linguistics}}\
  (\BPGS\ 1030--1039).
\newblock
\APACaddressPublisher{}{Association for Computational Linguistics}.
\PrintBackRefs{\CurrentBib}

\bibitem [\protect \citeauthoryear {%
Hammarstr\"om%
, Forkel%
, Haspelmath%
\BCBL {}\ \BBA {} Bank%
}{%
Hammarstr\"om%
\ \protect \BOthers {.}}{%
{\protect \APACyear {2016}}%
}]{%
glottolog2_7}
\APACinsertmetastar {%
glottolog2_7}%
\begin{APACrefauthors}%
Hammarstr\"om, H.%
, Forkel, R.%
, Haspelmath, M.%
\BCBL {}\ \BBA {} Bank, S.%
\end{APACrefauthors}%
\unskip\
\newblock
\APACrefYear{2016}.
\newblock
\APACrefbtitle {Glottolog 2.7} {Glottolog 2.7}.
\newblock
\APACaddressPublisher{Jena}{Max Planck Institute for the Science of Human
  History}.
\newblock
\APACrefnote{Available online at http://glottolog.org, Accessed on 2017-01-29}
\PrintBackRefs{\CurrentBib}

\bibitem [\protect \citeauthoryear {%
Haspelmath%
, Dryer%
, Gil%
\BCBL {}\ \BBA {} Comrie%
}{%
Haspelmath%
\ \protect \BOthers {.}}{%
{\protect \APACyear {2008}}%
}]{%
wals}
\APACinsertmetastar {%
wals}%
\begin{APACrefauthors}%
Haspelmath, M.%
, Dryer, M\BPBI S.%
, Gil, D.%
\BCBL {}\ \BBA {} Comrie, B.%
\end{APACrefauthors}%
\unskip\
\newblock
\APACrefYearMonthDay{2008}{}{}.
\newblock
\APACrefbtitle {The {World} {Atlas} of {Language} {Structures} Online.} {The
  {World} {Atlas} of {Language} {Structures} online.}
\newblock
\APAChowpublished {Max Planck Digital Library, Munich}.
\newblock
\APACrefnote{{http://wals.info/}}
\PrintBackRefs{\CurrentBib}

\bibitem [\protect \citeauthoryear {%
Hauer%
\ \BBA {} Kondrak%
}{%
Hauer%
\ \BBA {} Kondrak%
}{%
{\protect \APACyear {2011}}%
}]{%
Hauer2011}
\APACinsertmetastar {%
Hauer2011}%
\begin{APACrefauthors}%
Hauer, B.%
\BCBT {}\ \BBA {} Kondrak, G.%
\end{APACrefauthors}%
\unskip\
\newblock
\APACrefYearMonthDay{2011}{}{}.
\newblock
{\BBOQ}\APACrefatitle {Clustering semantically equivalent words into cognate
  sets in multilingual lists} {Clustering semantically equivalent words into
  cognate sets in multilingual lists}.{\BBCQ}
\newblock
\BIn{} \APACrefbtitle {Proceedings of the 5th International Joint {NLP}
  conference} {Proceedings of the 5th international joint {NLP} conference}\
  (\BPG~865-873).
\PrintBackRefs{\CurrentBib}

\bibitem [\protect \citeauthoryear {%
Herman%
}{%
Herman%
}{%
{\protect \APACyear {2000}}%
}]{%
hermanVulgarLatin}
\APACinsertmetastar {%
hermanVulgarLatin}%
\begin{APACrefauthors}%
Herman, J.%
\end{APACrefauthors}%
\unskip\
\newblock
\APACrefYear{2000}.
\newblock
\APACrefbtitle {Vulgar {Latin}} {Vulgar {Latin}}.
\newblock
\APACaddressPublisher{University Park, PA}{The Pennsylvania State University
  Press}.
\PrintBackRefs{\CurrentBib}

\bibitem [\protect \citeauthoryear {%
Hilpert%
\ \BBA {} Gries%
}{%
Hilpert%
\ \BBA {} Gries%
}{%
{\protect \APACyear {2016}}%
}]{%
hilpertGries2016}
\APACinsertmetastar {%
hilpertGries2016}%
\begin{APACrefauthors}%
Hilpert, M.%
\BCBT {}\ \BBA {} Gries, S\BPBI T.%
\end{APACrefauthors}%
\unskip\
\newblock
\APACrefYearMonthDay{2016}{}{}.
\newblock
{\BBOQ}\APACrefatitle {Quantitative approaches to diachronic corpus
  linguistics} {Quantitative approaches to diachronic corpus
  linguistics}.{\BBCQ}
\newblock
\BIn{} M.~Kyt{\"o}\ \BBA {} P.~Pahta\ (\BEDS), \APACrefbtitle {The {Cambridge}
  Handbook of {English} Historical Linguistics} {The {Cambridge} handbook of
  {English} historical linguistics}\ (\BPGS\ 36--53).
\newblock
\APACaddressPublisher{}{Cambridge University Press}.
\PrintBackRefs{\CurrentBib}

\bibitem [\protect \citeauthoryear {%
Hogeweg%
\ \BBA {} Hesper%
}{%
Hogeweg%
\ \BBA {} Hesper%
}{%
{\protect \APACyear {1984}}%
}]{%
progressiveAlignment}
\APACinsertmetastar {%
progressiveAlignment}%
\begin{APACrefauthors}%
Hogeweg, P.%
\BCBT {}\ \BBA {} Hesper, B.%
\end{APACrefauthors}%
\unskip\
\newblock
\APACrefYearMonthDay{1984}{}{}.
\newblock
{\BBOQ}\APACrefatitle {The alignment of sets of sequences and the construction
  of phyletic trees: an integrated method} {The alignment of sets of sequences
  and the construction of phyletic trees: an integrated method}.{\BBCQ}
\newblock
\APACjournalVolNumPages{Journal of molecular evolution}{20}{2}{175--186}.
\PrintBackRefs{\CurrentBib}

\bibitem [\protect \citeauthoryear {%
Hruschka%
\ \protect \BOthers {.}}{%
Hruschka%
\ \protect \BOthers {.}}{%
{\protect \APACyear {2015}}%
}]{%
hruschkaetal15}
\APACinsertmetastar {%
hruschkaetal15}%
\begin{APACrefauthors}%
Hruschka, D\BPBI J.%
, Branford, S.%
, Smitch, E\BPBI D.%
, Wilkins, J.%
, Meade, A.%
, Pagel, M.%
\BCBL {}\ \BBA {} Bhattachary, T.%
\end{APACrefauthors}%
\unskip\
\newblock
\APACrefYearMonthDay{2015}{}{}.
\newblock
{\BBOQ}\APACrefatitle {Detecting Regular Sound Changes in Linguistics as Events
  of Concerted Evolution} {Detecting regular sound changes in linguistics as
  events of concerted evolution}.{\BBCQ}
\newblock
\APACjournalVolNumPages{Current Biology}{25}{1}{1--9}.
\PrintBackRefs{\CurrentBib}

\bibitem [\protect \citeauthoryear {%
J\"ager%
}{%
J\"ager%
}{%
{\protect \APACyear {2013}}%
}]{%
jaeger13ldc}
\APACinsertmetastar {%
jaeger13ldc}%
\begin{APACrefauthors}%
J\"ager, G.%
\end{APACrefauthors}%
\unskip\
\newblock
\APACrefYearMonthDay{2013}{}{}.
\newblock
{\BBOQ}\APACrefatitle {Phylogenetic inference from word lists using weighted
  alignment with empirically determined weights} {Phylogenetic inference from
  word lists using weighted alignment with empirically determined
  weights}.{\BBCQ}
\newblock
\APACjournalVolNumPages{Language Dynamics and Change}{3}{2}{245--291}.
\PrintBackRefs{\CurrentBib}

\bibitem [\protect \citeauthoryear {%
J\"ager%
\ \BBA {} List%
}{%
J\"ager%
\ \BBA {} List%
}{%
{\protect \APACyear {2017}}%
}]{%
jaegerList17asr}
\APACinsertmetastar {%
jaegerList17asr}%
\begin{APACrefauthors}%
J\"ager, G.%
\BCBT {}\ \BBA {} List, J\BHBI M.%
\end{APACrefauthors}%
\unskip\
\newblock
\APACrefYearMonthDay{2017}{}{}.
\newblock
\APACrefbtitle {Using Ancestral State Reconstruction Methods for
  Onomasiological Reconstruction in Multilingual Word Lists.} {Using ancestral
  state reconstruction methods for onomasiological reconstruction in
  multilingual word lists.}
\newblock
\APACrefnote{Manuscript, T\"ubingen and Jena}
\PrintBackRefs{\CurrentBib}

\bibitem [\protect \citeauthoryear {%
J\"ager%
, List%
\BCBL {}\ \BBA {} Sofroniev%
}{%
J\"ager%
\ \protect \BOthers {.}}{%
{\protect \APACyear {2017}}%
}]{%
jaegerListSofroniev17}
\APACinsertmetastar {%
jaegerListSofroniev17}%
\begin{APACrefauthors}%
J\"ager, G.%
, List, J\BHBI M.%
\BCBL {}\ \BBA {} Sofroniev, P.%
\end{APACrefauthors}%
\unskip\
\newblock
\APACrefYearMonthDay{2017}{}{}.
\newblock
{\BBOQ}\APACrefatitle {Using support vector machines and state-of-the-art
  algorithms for phonetic alignment to identify cognates in multi-lingual
  wordlists} {Using support vector machines and state-of-the-art algorithms for
  phonetic alignment to identify cognates in multi-lingual wordlists}.{\BBCQ}
\newblock
\BIn{} \APACrefbtitle {Proceedings of the 15th Conference of the {European}
  Chapter of the {Association} for {Computational} {Linguistics}.} {Proceedings
  of the 15th conference of the {European} chapter of the {Association} for
  {Computational} {Linguistics}.}
\newblock
\APACaddressPublisher{}{ACL}.
\PrintBackRefs{\CurrentBib}

\bibitem [\protect \citeauthoryear {%
J\"ager%
\ \BBA {} Sofroniev%
}{%
J\"ager%
\ \BBA {} Sofroniev%
}{%
{\protect \APACyear {2016}}%
}]{%
jaegerSofroniev16Konvens}
\APACinsertmetastar {%
jaegerSofroniev16Konvens}%
\begin{APACrefauthors}%
J\"ager, G.%
\BCBT {}\ \BBA {} Sofroniev, P.%
\end{APACrefauthors}%
\unskip\
\newblock
\APACrefYearMonthDay{2016}{}{}.
\newblock
{\BBOQ}\APACrefatitle {Automatic cognate classification with a {Support}
  {Vector} {Machine}} {Automatic cognate classification with a {Support}
  {Vector} {Machine}}.{\BBCQ}
\newblock
\BIn{} S.~Dipper, F.~Neubarth\BCBL {}\ \BBA {} H.~Zinsmeister\ (\BEDS),
  \APACrefbtitle {Proceedings of the 13th {Conference} on {Natural} {Language}
  {Processing}} {Proceedings of the 13th {Conference} on {Natural} {Language}
  {Processing}}\ (\BVOL~16, \BPGS\ 128--134).
\PrintBackRefs{\CurrentBib}

\bibitem [\protect \citeauthoryear {%
Kay%
}{%
Kay%
}{%
{\protect \APACyear {1964}}%
}]{%
kay1964}
\APACinsertmetastar {%
kay1964}%
\begin{APACrefauthors}%
Kay, M.%
\end{APACrefauthors}%
\unskip\
\newblock
\APACrefYear{1964}.
\newblock
\APACrefbtitle {The logic of cognate recognition in historical linguistics}
  {The logic of cognate recognition in historical linguistics}.
\newblock
\APACaddressPublisher{}{Rand Corporation}.
\PrintBackRefs{\CurrentBib}

\bibitem [\protect \citeauthoryear {%
Kessler%
}{%
Kessler%
}{%
{\protect \APACyear {2001}}%
}]{%
kessler2001}
\APACinsertmetastar {%
kessler2001}%
\begin{APACrefauthors}%
Kessler, B.%
\end{APACrefauthors}%
\unskip\
\newblock
\APACrefYear{2001}.
\newblock
\APACrefbtitle {The significance of word lists} {The significance of word
  lists}.
\newblock
\APACaddressPublisher{Stanford}{CSLI Publications}.
\PrintBackRefs{\CurrentBib}

\bibitem [\protect \citeauthoryear {%
Kondrak%
}{%
Kondrak%
}{%
{\protect \APACyear {2002}}%
}]{%
kondrak02}
\APACinsertmetastar {%
kondrak02}%
\begin{APACrefauthors}%
Kondrak, G.%
\end{APACrefauthors}%
\unskip\
\newblock
\APACrefYear{2002}.
\unskip\
\newblock
\APACrefbtitle {Algorithms for Language Reconstruction} {Algorithms for
  language reconstruction}\ \APACtypeAddressSchool {\BUPhD}{}{}.
\unskip\
\newblock
\APACaddressSchool {}{University of Toronto}.
\PrintBackRefs{\CurrentBib}

\bibitem [\protect \citeauthoryear {%
Kooperberg%
}{%
Kooperberg%
}{%
{\protect \APACyear {2016}}%
}]{%
logspline}
\APACinsertmetastar {%
logspline}%
\begin{APACrefauthors}%
Kooperberg, C.%
\end{APACrefauthors}%
\unskip\
\newblock
\APACrefYearMonthDay{2016}{}{}.
\newblock
\APACrefbtitle {Package `logspline'.} {Package `logspline'.}
\newblock
\APAChowpublished
  {https://cran.r-project.org/web/packages/logspline/index.html}.
\newblock
\APACrefnote{version 2.1.9}
\PrintBackRefs{\CurrentBib}

\bibitem [\protect \citeauthoryear {%
Kroonen%
}{%
Kroonen%
}{%
{\protect \APACyear {2013}}%
}]{%
kroonen12}
\APACinsertmetastar {%
kroonen12}%
\begin{APACrefauthors}%
Kroonen, G.%
\end{APACrefauthors}%
\unskip\
\newblock
\APACrefYear{2013}.
\newblock
\APACrefbtitle {Etymological Dictionary of {Proto-Germanic}} {Etymological
  dictionary of {Proto-Germanic}}.
\newblock
\APACaddressPublisher{Leiden, Boston}{Brill}.
\PrintBackRefs{\CurrentBib}

\bibitem [\protect \citeauthoryear {%
Lewis%
, Simons%
\BCBL {}\ \BBA {} Fennig%
}{%
Lewis%
\ \protect \BOthers {.}}{%
{\protect \APACyear {2016}}%
}]{%
ethnologue2016}
\APACinsertmetastar {%
ethnologue2016}%
\begin{APACrefauthors}%
Lewis, M\BPBI P.%
, Simons, G\BPBI F.%
\BCBL {}\ \BBA {} Fennig, C\BPBI D.%
\end{APACrefauthors}%
\ (\BEDS).
\unskip\
\newblock
\APACrefYear{2016}.
\newblock
\APACrefbtitle {Ethnologue: Languages of the World} {Ethnologue: Languages of
  the world}\ (\PrintOrdinal{Nineteenth}\ \BEd).
\newblock
\APACaddressPublisher{Dallas, Texas}{SIL International}.
\PrintBackRefs{\CurrentBib}

\bibitem [\protect \citeauthoryear {%
List%
}{%
List%
}{%
{\protect \APACyear {2012}}%
}]{%
list2012lexstat}
\APACinsertmetastar {%
list2012lexstat}%
\begin{APACrefauthors}%
List, J\BHBI M.%
\end{APACrefauthors}%
\unskip\
\newblock
\APACrefYearMonthDay{2012}{}{}.
\newblock
{\BBOQ}\APACrefatitle {LexStat: Automatic detection of cognates in multilingual
  wordlists} {Lexstat: Automatic detection of cognates in multilingual
  wordlists}.{\BBCQ}
\newblock
\BIn{} M.~Butt\ \BBA {} J.~Proki{\'c}\ (\BEDS), \APACrefbtitle {Proceedings of
  LINGVIS \& UNCLH, Workshop at EACL 2012} {Proceedings of lingvis \& unclh,
  workshop at eacl 2012}\ (\BPGS\ 117--125).
\newblock
\APACaddressPublisher{Avignon}{}.
\PrintBackRefs{\CurrentBib}

\bibitem [\protect \citeauthoryear {%
List%
}{%
List%
}{%
{\protect \APACyear {2014}}%
}]{%
list14}
\APACinsertmetastar {%
list14}%
\begin{APACrefauthors}%
List, J\BHBI M.%
\end{APACrefauthors}%
\unskip\
\newblock
\APACrefYear{2014}.
\newblock
\APACrefbtitle {Sequence Comparison in Historical Linguistics} {Sequence
  comparison in historical linguistics}.
\newblock
\APACaddressPublisher{D\"usseldorf}{D\"usseldorf University Press}.
\PrintBackRefs{\CurrentBib}

\bibitem [\protect \citeauthoryear {%
Lowe%
\ \BBA {} Mazaudon%
}{%
Lowe%
\ \BBA {} Mazaudon%
}{%
{\protect \APACyear {1994}}%
}]{%
loweMazaudon1994}
\APACinsertmetastar {%
loweMazaudon1994}%
\begin{APACrefauthors}%
Lowe, J\BPBI B.%
\BCBT {}\ \BBA {} Mazaudon, M.%
\end{APACrefauthors}%
\unskip\
\newblock
\APACrefYearMonthDay{1994}{}{}.
\newblock
{\BBOQ}\APACrefatitle {The reconstruction engine: a computer implementation of
  the comparative method} {The reconstruction engine: a computer implementation
  of the comparative method}.{\BBCQ}
\newblock
\APACjournalVolNumPages{Computational Linguistics}{20}{3}{381--417}.
\PrintBackRefs{\CurrentBib}

\bibitem [\protect \citeauthoryear {%
MacMahon%
\ \BBA {} MacMahon%
}{%
MacMahon%
\ \BBA {} MacMahon%
}{%
{\protect \APACyear {2006}}%
}]{%
macmahonMacmahon2006}
\APACinsertmetastar {%
macmahonMacmahon2006}%
\begin{APACrefauthors}%
MacMahon, A.%
\BCBT {}\ \BBA {} MacMahon, R.%
\end{APACrefauthors}%
\unskip\
\newblock
\APACrefYearMonthDay{2006}{}{}.
\newblock
{\BBOQ}\APACrefatitle {Why linguists don’t do dates: evidence from
  {Indo-European} and {Australian} languages} {Why linguists don’t do dates:
  evidence from {Indo-European} and {Australian} languages}.{\BBCQ}
\newblock
\BIn{} P.~Forster\ \BBA {} C.~Renfrew\ (\BEDS), \APACrefbtitle {Phylogenetic
  methods and the prehistory of languages} {Phylogenetic methods and the
  prehistory of languages}\ (\BPGS\ 153--160).
\newblock
\APACaddressPublisher{Cambridge, UK}{McDonald Institute for Archaeological
  Research, Cambridge}.
\PrintBackRefs{\CurrentBib}

\bibitem [\protect \citeauthoryear {%
McMahon%
\ \BBA {} McMahon%
}{%
McMahon%
\ \BBA {} McMahon%
}{%
{\protect \APACyear {2005}}%
}]{%
mcmahonMcmahon2005}
\APACinsertmetastar {%
mcmahonMcmahon2005}%
\begin{APACrefauthors}%
McMahon, A.%
\BCBT {}\ \BBA {} McMahon, R.%
\end{APACrefauthors}%
\unskip\
\newblock
\APACrefYear{2005}.
\newblock
\APACrefbtitle {Language Classification by Numbers} {Language classification by
  numbers}.
\newblock
\APACaddressPublisher{Oxford}{Oxford University Press}.
\PrintBackRefs{\CurrentBib}

\bibitem [\protect \citeauthoryear {%
Meillet%
}{%
Meillet%
}{%
{\protect \APACyear {1954}}%
}]{%
meillet1925}
\APACinsertmetastar {%
meillet1925}%
\begin{APACrefauthors}%
Meillet, A.%
\end{APACrefauthors}%
\unskip\
\newblock
\APACrefYear{1954}.
\newblock
\APACrefbtitle {{La méthode comparative en linguistique historique
  \textnormal{[The comparative method in historical linguistics]}}} {{La
  méthode comparative en linguistique historique \textnormal{[The comparative
  method in historical linguistics]}}}.
\newblock
\APACaddressPublisher{Paris}{Honoré Champion}.
\newblock
\APACrefnote{reprint}
\PrintBackRefs{\CurrentBib}

\bibitem [\protect \citeauthoryear {%
Meyer-L{\"u}bke%
}{%
Meyer-L{\"u}bke%
}{%
{\protect \APACyear {1935}}%
}]{%
meyerLuebke}
\APACinsertmetastar {%
meyerLuebke}%
\begin{APACrefauthors}%
Meyer-L{\"u}bke, W.%
\end{APACrefauthors}%
\unskip\
\newblock
\APACrefYear{1935}.
\newblock
\APACrefbtitle {Romanisches etymologisches {W{\"o}rterbuch}} {Romanisches
  etymologisches {W{\"o}rterbuch}}.
\newblock
\APACaddressPublisher{Heidelberg}{Carl Winters Universitätsbuchhandlung}.
\newblock
\APACrefnote{3.\ {Auflage}}
\PrintBackRefs{\CurrentBib}

\bibitem [\protect \citeauthoryear {%
Needleman%
\ \BBA {} Wunsch%
}{%
Needleman%
\ \BBA {} Wunsch%
}{%
{\protect \APACyear {1970}}%
}]{%
needlemanWunsch}
\APACinsertmetastar {%
needlemanWunsch}%
\begin{APACrefauthors}%
Needleman, S\BPBI B.%
\BCBT {}\ \BBA {} Wunsch, C\BPBI D.%
\end{APACrefauthors}%
\unskip\
\newblock
\APACrefYearMonthDay{1970}{}{}.
\newblock
{\BBOQ}\APACrefatitle {A general method applicable to the search for
  similarities in the amino acid sequence of two proteins} {A general method
  applicable to the search for similarities in the amino acid sequence of two
  proteins}.{\BBCQ}
\newblock
\APACjournalVolNumPages{Journal of Molecular Biology}{48}{}{443–453}.
\PrintBackRefs{\CurrentBib}

\bibitem [\protect \citeauthoryear {%
Nguyen%
, Schmidt%
, von Haeseler%
\BCBL {}\ \BBA {} Minh%
}{%
Nguyen%
\ \protect \BOthers {.}}{%
{\protect \APACyear {2015}}%
}]{%
iqtree}
\APACinsertmetastar {%
iqtree}%
\begin{APACrefauthors}%
Nguyen, L\BHBI T.%
, Schmidt, H\BPBI A.%
, von Haeseler, A.%
\BCBL {}\ \BBA {} Minh, B\BPBI Q.%
\end{APACrefauthors}%
\unskip\
\newblock
\APACrefYearMonthDay{2015}{}{}.
\newblock
{\BBOQ}\APACrefatitle {{IQ-TREE}: a fast and effective stochastic algorithm for
  estimating maximum-likelihood phylogenies} {{IQ-TREE}: a fast and effective
  stochastic algorithm for estimating maximum-likelihood phylogenies}.{\BBCQ}
\newblock
\APACjournalVolNumPages{Molecular biology and evolution}{32}{1}{268--274}.
\PrintBackRefs{\CurrentBib}

\bibitem [\protect \citeauthoryear {%
Notredame%
, Higgins%
\BCBL {}\ \BBA {} Heringa%
}{%
Notredame%
\ \protect \BOthers {.}}{%
{\protect \APACyear {2000}}%
}]{%
tCoffee}
\APACinsertmetastar {%
tCoffee}%
\begin{APACrefauthors}%
Notredame, C.%
, Higgins, D\BPBI G.%
\BCBL {}\ \BBA {} Heringa, J.%
\end{APACrefauthors}%
\unskip\
\newblock
\APACrefYearMonthDay{2000}{}{}.
\newblock
{\BBOQ}\APACrefatitle {T-{Coffee}: A novel method for fast and accurate
  multiple sequence alignment} {T-{Coffee}: A novel method for fast and
  accurate multiple sequence alignment}.{\BBCQ}
\newblock
\APACjournalVolNumPages{Journal of molecular biology}{302}{1}{205--217}.
\PrintBackRefs{\CurrentBib}

\bibitem [\protect \citeauthoryear {%
Oakes%
}{%
Oakes%
}{%
{\protect \APACyear {2000}}%
}]{%
oakes2000}
\APACinsertmetastar {%
oakes2000}%
\begin{APACrefauthors}%
Oakes, M\BPBI P.%
\end{APACrefauthors}%
\unskip\
\newblock
\APACrefYearMonthDay{2000}{}{}.
\newblock
{\BBOQ}\APACrefatitle {Computer estimation of vocabulary in a protolanguage
  from word lists in four daughter languages} {Computer estimation of
  vocabulary in a protolanguage from word lists in four daughter
  languages}.{\BBCQ}
\newblock
\APACjournalVolNumPages{Journal of Quantitative Linguistics}{7}{3}{233--243}.
\PrintBackRefs{\CurrentBib}

\bibitem [\protect \citeauthoryear {%
Pagel%
, Atkinson%
, Calude%
\BCBL {}\ \BBA {} Meade%
}{%
Pagel%
\ \protect \BOthers {.}}{%
{\protect \APACyear {2013}}%
}]{%
pagel2013}
\APACinsertmetastar {%
pagel2013}%
\begin{APACrefauthors}%
Pagel, M.%
, Atkinson, Q\BPBI D.%
, Calude, A\BPBI S.%
\BCBL {}\ \BBA {} Meade, A.%
\end{APACrefauthors}%
\unskip\
\newblock
\APACrefYearMonthDay{2013}{}{}.
\newblock
{\BBOQ}\APACrefatitle {Ultraconserved words point to deep language ancestry
  across {Eurasia}} {Ultraconserved words point to deep language ancestry
  across {Eurasia}}.{\BBCQ}
\newblock
\APACjournalVolNumPages{Proceedings of the National Academy of
  Sciences}{110}{21}{8471--8476}.
\PrintBackRefs{\CurrentBib}

\bibitem [\protect \citeauthoryear {%
Pagel%
, Atkinson%
\BCBL {}\ \BBA {} Meade%
}{%
Pagel%
\ \protect \BOthers {.}}{%
{\protect \APACyear {2007}}%
}]{%
pageletal2007}
\APACinsertmetastar {%
pageletal2007}%
\begin{APACrefauthors}%
Pagel, M.%
, Atkinson, Q\BPBI D.%
\BCBL {}\ \BBA {} Meade, A.%
\end{APACrefauthors}%
\unskip\
\newblock
\APACrefYearMonthDay{2007}{}{}.
\newblock
{\BBOQ}\APACrefatitle {Frequency of word-use predicts rates of lexical
  evolution throughout {Indo-European} history} {Frequency of word-use predicts
  rates of lexical evolution throughout {Indo-European} history}.{\BBCQ}
\newblock
\APACjournalVolNumPages{Nature}{449}{7163}{717--720}.
\PrintBackRefs{\CurrentBib}

\bibitem [\protect \citeauthoryear {%
Pietrusewsky%
}{%
Pietrusewsky%
}{%
{\protect \APACyear {2008}}%
}]{%
pietrusewsky2008}
\APACinsertmetastar {%
pietrusewsky2008}%
\begin{APACrefauthors}%
Pietrusewsky, M.%
\end{APACrefauthors}%
\unskip\
\newblock
\APACrefYearMonthDay{2008}{}{}.
\newblock
{\BBOQ}\APACrefatitle {Craniometric variation in {Southeast} {Asia} and
  neighboring regions: a multivariate analysis of cranial measurements}
  {Craniometric variation in {Southeast} {Asia} and neighboring regions: a
  multivariate analysis of cranial measurements}.{\BBCQ}
\newblock
\APACjournalVolNumPages{Human evolution}{23}{1--2}{49--86}.
\PrintBackRefs{\CurrentBib}

\bibitem [\protect \citeauthoryear {%
Raghavan%
, Albert%
\BCBL {}\ \BBA {} Kumara%
}{%
Raghavan%
\ \protect \BOthers {.}}{%
{\protect \APACyear {2007}}%
}]{%
labelPropagation}
\APACinsertmetastar {%
labelPropagation}%
\begin{APACrefauthors}%
Raghavan, U\BPBI N.%
, Albert, R.%
\BCBL {}\ \BBA {} Kumara, S.%
\end{APACrefauthors}%
\unskip\
\newblock
\APACrefYearMonthDay{2007}{}{}.
\newblock
{\BBOQ}\APACrefatitle {Near linear time algorithm to detect community
  structures in large-scale networks} {Near linear time algorithm to detect
  community structures in large-scale networks}.{\BBCQ}
\newblock
\APACjournalVolNumPages{Physical Review E}{76}{3}{036106}.
\PrintBackRefs{\CurrentBib}

\bibitem [\protect \citeauthoryear {%
Rama%
}{%
Rama%
}{%
{\protect \APACyear {2013}}%
}]{%
rama13}
\APACinsertmetastar {%
rama13}%
\begin{APACrefauthors}%
Rama, T.%
\end{APACrefauthors}%
\unskip\
\newblock
\APACrefYearMonthDay{2013}{}{}.
\newblock
{\BBOQ}\APACrefatitle {Phonotactic diversity predicts the time depth of the
  world’s language families} {Phonotactic diversity predicts the time depth
  of the world’s language families}.{\BBCQ}
\newblock
\APACjournalVolNumPages{PLoS ONE}{8}{5}{e63238}.
\PrintBackRefs{\CurrentBib}

\bibitem [\protect \citeauthoryear {%
Rama%
}{%
Rama%
}{%
{\protect \APACyear {2015}}%
}]{%
rama2015}
\APACinsertmetastar {%
rama2015}%
\begin{APACrefauthors}%
Rama, T.%
\end{APACrefauthors}%
\unskip\
\newblock
\APACrefYearMonthDay{2015}{}{}.
\newblock
{\BBOQ}\APACrefatitle {Automatic cognate identification with gap-weighted
  string subsequences} {Automatic cognate identification with gap-weighted
  string subsequences}.{\BBCQ}
\newblock
\BIn{} \APACrefbtitle {Proceedings of the {North} {American} {Association} for
  {Computational} {Linguistics}} {Proceedings of the {North} {American}
  {Association} for {Computational} {Linguistics}}\ (\BPGS\ 1227--1231).
\newblock
\APACaddressPublisher{}{Association for Computational Linguistics}.
\PrintBackRefs{\CurrentBib}

\bibitem [\protect \citeauthoryear {%
Renfrew%
}{%
Renfrew%
}{%
{\protect \APACyear {1987}}%
}]{%
renfrew1987}
\APACinsertmetastar {%
renfrew1987}%
\begin{APACrefauthors}%
Renfrew, C.%
\end{APACrefauthors}%
\unskip\
\newblock
\APACrefYear{1987}.
\newblock
\APACrefbtitle {Archaeology and language: the puzzle of {Indo-European}
  origins} {Archaeology and language: the puzzle of {Indo-European} origins}.
\newblock
\APACaddressPublisher{Cambridge, UK}{Cambridge University Press}.
\PrintBackRefs{\CurrentBib}

\bibitem [\protect \citeauthoryear {%
Ringe%
}{%
Ringe%
}{%
{\protect \APACyear {1992}}%
}]{%
ringe1992}
\APACinsertmetastar {%
ringe1992}%
\begin{APACrefauthors}%
Ringe, D\BPBI A.%
\end{APACrefauthors}%
\unskip\
\newblock
\APACrefYearMonthDay{1992}{}{}.
\newblock
{\BBOQ}\APACrefatitle {On calculating the factor of chance in language
  comparison} {On calculating the factor of chance in language
  comparison}.{\BBCQ}
\newblock
\APACjournalVolNumPages{Transactions of the American Philosophical
  Society}{82}{1}{1--110}.
\PrintBackRefs{\CurrentBib}

\bibitem [\protect \citeauthoryear {%
Ringe%
, Warnow%
\BCBL {}\ \BBA {} Taylor%
}{%
Ringe%
\ \protect \BOthers {.}}{%
{\protect \APACyear {2002}}%
}]{%
ringeWarnowTaylor2002}
\APACinsertmetastar {%
ringeWarnowTaylor2002}%
\begin{APACrefauthors}%
Ringe, D\BPBI A.%
, Warnow, T.%
\BCBL {}\ \BBA {} Taylor, A.%
\end{APACrefauthors}%
\unskip\
\newblock
\APACrefYearMonthDay{2002}{}{}.
\newblock
{\BBOQ}\APACrefatitle {Indo-{European} and computational cladistics}
  {Indo-{European} and computational cladistics}.{\BBCQ}
\newblock
\APACjournalVolNumPages{Transactions of the Philological
  Society}{100}{1}{59--129}.
\PrintBackRefs{\CurrentBib}

\bibitem [\protect \citeauthoryear {%
Ronquist%
\ \BBA {} Huelsenbeck%
}{%
Ronquist%
\ \BBA {} Huelsenbeck%
}{%
{\protect \APACyear {2003}}%
}]{%
mrbayes3}
\APACinsertmetastar {%
mrbayes3}%
\begin{APACrefauthors}%
Ronquist, F.%
\BCBT {}\ \BBA {} Huelsenbeck, J\BPBI P.%
\end{APACrefauthors}%
\unskip\
\newblock
\APACrefYearMonthDay{2003}{}{}.
\newblock
{\BBOQ}\APACrefatitle {{MrBayes} 3: Bayesian phylogenetic inference under mixed
  models} {{MrBayes} 3: Bayesian phylogenetic inference under mixed
  models}.{\BBCQ}
\newblock
\APACjournalVolNumPages{Bioinformatics}{19}{12}{1572--1574}.
\PrintBackRefs{\CurrentBib}

\bibitem [\protect \citeauthoryear {%
Ross%
\ \BBA {} Durie%
}{%
Ross%
\ \BBA {} Durie%
}{%
{\protect \APACyear {1996}}%
}]{%
rossDurie96}
\APACinsertmetastar {%
rossDurie96}%
\begin{APACrefauthors}%
Ross, M.%
\BCBT {}\ \BBA {} Durie, M.%
\end{APACrefauthors}%
\unskip\
\newblock
\APACrefYearMonthDay{1996}{}{}.
\newblock
{\BBOQ}\APACrefatitle {Introduction} {Introduction}.{\BBCQ}
\newblock
\BIn{} M.~Durie\ \BBA {} M.~Ross\ (\BEDS), \APACrefbtitle {The Comparative
  Method Reviewed. Regularity and Irregularity in Language Change} {The
  comparative method reviewed. regularity and irregularity in language change}\
  (\BPGS\ 3--38).
\newblock
\APACaddressPublisher{New York and Oxford}{Oxford University Press}.
\PrintBackRefs{\CurrentBib}

\bibitem [\protect \citeauthoryear {%
Saitou%
\ \BBA {} Nei%
}{%
Saitou%
\ \BBA {} Nei%
}{%
{\protect \APACyear {1987}}%
}]{%
saitouNei87}
\APACinsertmetastar {%
saitouNei87}%
\begin{APACrefauthors}%
Saitou, N.%
\BCBT {}\ \BBA {} Nei, M.%
\end{APACrefauthors}%
\unskip\
\newblock
\APACrefYearMonthDay{1987}{}{}.
\newblock
{\BBOQ}\APACrefatitle {The neighbor-joining method: a new method for
  reconstructing phylogenetic trees.} {The neighbor-joining method: a new
  method for reconstructing phylogenetic trees.}{\BBCQ}
\newblock
\APACjournalVolNumPages{Molecular biology and evolution}{4}{4}{406--425}.
\PrintBackRefs{\CurrentBib}

\bibitem [\protect \citeauthoryear {%
Stamatakis%
}{%
Stamatakis%
}{%
{\protect \APACyear {2014}}%
}]{%
raxml8}
\APACinsertmetastar {%
raxml8}%
\begin{APACrefauthors}%
Stamatakis, A.%
\end{APACrefauthors}%
\unskip\
\newblock
\APACrefYearMonthDay{2014}{}{}.
\newblock
{\BBOQ}\APACrefatitle {{RAxML} version 8: a tool for phylogenetic analysis and
  post-analysis of large phylogenies} {{RAxML} version 8: a tool for
  phylogenetic analysis and post-analysis of large phylogenies}.{\BBCQ}
\newblock
\APACjournalVolNumPages{Bioinformatics}{30}{9}{1312--1313}.
\PrintBackRefs{\CurrentBib}

\bibitem [\protect \citeauthoryear {%
Swadesh%
}{%
Swadesh%
}{%
{\protect \APACyear {1952}}%
}]{%
swadesh52}
\APACinsertmetastar {%
swadesh52}%
\begin{APACrefauthors}%
Swadesh, M.%
\end{APACrefauthors}%
\unskip\
\newblock
\APACrefYearMonthDay{1952}{}{}.
\newblock
{\BBOQ}\APACrefatitle {Lexico-statistic dating of prehistoric ethnic contacts}
  {Lexico-statistic dating of prehistoric ethnic contacts}.{\BBCQ}
\newblock
\APACjournalVolNumPages{Proceedings of the American Philosophical
  Society}{96}{4}{452--463}.
\PrintBackRefs{\CurrentBib}

\bibitem [\protect \citeauthoryear {%
Swadesh%
}{%
Swadesh%
}{%
{\protect \APACyear {1955}}%
}]{%
swadesh55}
\APACinsertmetastar {%
swadesh55}%
\begin{APACrefauthors}%
Swadesh, M.%
\end{APACrefauthors}%
\unskip\
\newblock
\APACrefYearMonthDay{1955}{}{}.
\newblock
{\BBOQ}\APACrefatitle {Towards greater accuracy in lexicostatistic dating}
  {Towards greater accuracy in lexicostatistic dating}.{\BBCQ}
\newblock
\APACjournalVolNumPages{International Journal of American
  Linguistics}{21}{}{121--137}.
\PrintBackRefs{\CurrentBib}

\bibitem [\protect \citeauthoryear {%
Turchin%
, Peiros%
\BCBL {}\ \BBA {} Gell-Mann%
}{%
Turchin%
\ \protect \BOthers {.}}{%
{\protect \APACyear {2010}}%
}]{%
Turchin2010}
\APACinsertmetastar {%
Turchin2010}%
\begin{APACrefauthors}%
Turchin, P.%
, Peiros, I.%
\BCBL {}\ \BBA {} Gell-Mann, M.%
\end{APACrefauthors}%
\unskip\
\newblock
\APACrefYearMonthDay{2010}{}{}.
\newblock
{\BBOQ}\APACrefatitle {Analyzing genetic connections between languages by
  matching consonant classes} {Analyzing genetic connections between languages
  by matching consonant classes}.{\BBCQ}
\newblock
\APACjournalVolNumPages{Journal of Language Relationship}{3}{}{117-126}.
\PrintBackRefs{\CurrentBib}

\bibitem [\protect \citeauthoryear {%
Weiss%
}{%
Weiss%
}{%
{\protect \APACyear {2015}}%
}]{%
weiss2015}
\APACinsertmetastar {%
weiss2015}%
\begin{APACrefauthors}%
Weiss, M.%
\end{APACrefauthors}%
\unskip\
\newblock
\APACrefYearMonthDay{2015}{}{}.
\newblock
{\BBOQ}\APACrefatitle {The comparative method} {The comparative method}.{\BBCQ}
\newblock
\BIn{} C.~Bowern\ \BBA {} B.~Evans\ (\BEDS), \APACrefbtitle {The {Routledge}
  Handbook of Historical Linguistics} {The {Routledge} handbook of historical
  linguistics}\ (\BPGS\ 119--121).
\newblock
\APACaddressPublisher{}{Routledge}.
\PrintBackRefs{\CurrentBib}

\bibitem [\protect \citeauthoryear {%
Wichmann%
, Holman%
\BCBL {}\ \BBA {} Brown%
}{%
Wichmann%
\ \protect \BOthers {.}}{%
{\protect \APACyear {2016}}%
}]{%
asjp17}
\APACinsertmetastar {%
asjp17}%
\begin{APACrefauthors}%
Wichmann, S.%
, Holman, E\BPBI W.%
\BCBL {}\ \BBA {} Brown, C\BPBI H.%
\end{APACrefauthors}%
\unskip\
\newblock
\APACrefYearMonthDay{2016}{}{}.
\newblock
\APACrefbtitle {The {ASJP} Database (version 17).} {The {ASJP} database
  (version 17).}
\newblock
\APAChowpublished {http://asjp.clld.org/}.
\PrintBackRefs{\CurrentBib}

\end{thebibliography}
\end{document}